\documentclass[hidelinks]{article}

\usepackage[final]{neurips_2024}

\usepackage[utf8]{inputenc} 
\usepackage[T1]{fontenc}    
\usepackage{hyperref}       
\usepackage{url}            
\usepackage{booktabs}       
\usepackage{amsfonts}       
\usepackage{nicefrac}       
\usepackage{microtype}      
\usepackage{xcolor}         
\usepackage{macros}

\usepackage{upgreek}
\usepackage{fmtcount}

\title{Higher-Order Causal Message Passing for Experimentation with Complex Interference}

\author{Mohsen Bayati$^1$ \quad
  Yuwei Luo$^1$ \quad
  William Overman$^1$ \quad
  Sadegh Shirani$^1$ \quad
  Ruoxuan Xiong$^2$ \\ \\
  $^1$ Stanford Graduate School of Business \quad $^2$ Emory University \\
  {\small \texttt{\{bayati, yuweiluo, wpo, sshirani\}@stanford.edu}, \quad \texttt{ruoxuan.xiong@emory.edu}}
  }

\begin{document}

\maketitle

\begin{abstract}
Accurate estimation of treatment effects is essential for decision-making across various scientific fields. This task, however, becomes challenging in areas like social sciences and online marketplaces, where treating one experimental unit can influence outcomes for others through direct or indirect interactions. Such interference can lead to biased treatment effect estimates, particularly when the structure of these interactions is unknown. We address this challenge by introducing a new class of estimators based on causal message-passing, specifically designed for settings with pervasive, unknown interference. Our estimator draws on information from the sample mean and variance of unit outcomes and treatments over time, enabling efficient use of observed data to estimate the evolution of the system state. Concretely, we construct non-linear features from the moments of unit outcomes and treatments and then learn a function that maps these features to future mean and variance of unit outcomes. This allows for the estimation of the treatment effect over time. Extensive simulations across multiple domains, using synthetic and real network data, demonstrate the efficacy of our approach in estimating total treatment effect dynamics, even in cases where interference exhibits non-monotonic behavior in the probability of treatment. 
\end{abstract}

\section{Introduction}
\label{sec:introduction}

Randomized experiments are widely recognized as a reliable method in data-driven decision-making for determining the causal effects of new interventions, such as medical treatments or upgrades of market products. The conventional approach involves administering the new treatment to a randomly selected subset of the observation units (e.g., patients, products, or geographical areas), referred to as the \emph{treatment} group, and comparing their outcomes with those units who received no treatment, the \emph{control} group. However, the validity of these methods substantially relies on the assumption that treating a group of units does not interfere with the outcomes of the control units, known as the Stable Unit Treatment Value Assumption (SUTVA) \citep{ cox1958planning, rubin1978bayesian, manski1990nonparametric,imbens2015causal,sussman2017elements}.

In many social science and online marketplace scenarios, treating one unit impacts not only its outcome but also the outcomes of units that directly or indirectly interact with the treated unit \citep{bond201261, blake2014why, holtz2020reducing, johari2022experimental,bright2022reducing}. This \emph{interference} of treatments and outcomes makes estimating the causal effect of the treatment particularly challenging. Considering the network of interactions, when a unit is treated, its interactions with neighboring units lead to subsequent changes in their outcomes. These interactions continue over the experimental time horizon and may display complex behaviors. For example, as the treatment is expanded to a larger population, the interference effect may intensify or diminish. This necessitates efficient data usage and robust estimators to capture and adapt to such intricacies.

Given the complexity of analyzing interference phenomena, research on network interference often relies on a series of simplifying assumptions. One common assumption is to ignore variations over time and assume outcomes are observed at equilibrium, which discards valuable information before the system reaches equilibrium. To reduce the complexity of the analysis, further assumptions are imposed on the nature and level of interference \citep{choi2017estimation,cortez2022staggered, li2022network}, such as the neighborhood interference assumption or assumptions on the maximum degree of the network. Additionally, a frequently made assumption to help estimate treatment effects is that the interference network is observed \citep{chen2024optimized, agarwal2022network, jia2024clustered}, which is impractical in some settings, such as under pervasive interference. For example, in large-scale online platforms, units may interact through competing platforms, making it difficult to account for all sources of interference. Our aim in this paper is to relax these assumptions.

The impact of network interference can be intricate, particularly when considering interactions among units over time. For example, applying the treatment to one unit can have \emph{spillover effects} on some control units, or one unit’s outcome can directly exert \emph{peer effects} on other units’ outcomes. Simultaneously, treatments with long-lasting effects can have \emph{carryover effects} to future time periods, and units’ outcomes can be serially correlated or have \emph{autocorrelation} over time. Consequently, whenever SUTVA fails to hold, the number of potential outcomes grows exponentially with the population size and the time horizon of the experiment. This renders the estimation of causal effects under general interference structures impossible due to non-identifiability challenges \citep{manski2013identification,aronow2017estimating,basse2018limitations, karwa2018systematic,forastiere2022estimating}.

Recently, \cite{shirani2024causal} introduced a new framework called \emph{Causal Message-Passing} (CMP) to address the challenge of causal effect estimation under unobserved pervasive interference. Their methodology relies on observing outcomes over time and is rooted in statistical physics \citep{mezard1986spin,mezard2009information} and approximate message passing (AMP) \citep{donoho2009message,bayati2011dynamics} from high dimensional statistics. Instead of investigating the complex relationships among units, which requires knowledge of the network, CMP focuses on the dynamics of one-dimensional quantities, such as the sample mean and sample variance of units' outcomes over time. These one-dimensional equations, also known as \emph{state evolution} equations, can help track how the administered intervention propagates through the network of units over time, which enables the estimation of counterfactual scenarios. However, it remains underexplored how to use state evolution to estimate causal effects.

In this work, we propose to utilize machine learning to learn a mapping that updates key parameters of the distribution of outcomes over time for causal effect estimation. This is achieved by introducing a set of non-linear feature functions that act on the observed outcomes, creating a “basis” for the learning task. By training a properly designed machine learning model on this extracted basis, we estimate the Total Treatment Effect (TTE), also known as the Global Treatment Effect (GTE) or Global Average Treatment Effect (GATE), which measures the causal effect of altering the treatment scenario from treating no one to treating everyone. 
The result is a family of estimators that allow one to extract more information from the experimental data, thereby ensuring efficient use of the data.

To be more specific, this work builds on the foundation established by \cite{shirani2024causal}, extending their method in two directions by introducing Higher-Order Causal Message Passing (HO-CMP) algorithms. First, HO-CMP incorporates higher-order moments of unit outcomes, unlike \cite{shirani2024causal}'s approach, which only employs the first moments for estimation. Second, while \cite{shirani2024causal} focus solely on two-stage experiments with two different probabilities of treatment, our work leverages the additional data provided by having more than two experimental stages with multiple probabilities of treatment. Thus, our work aligns with the common practice in the tech industry of rolling out treatments through a sequence of experiments \citep{kohavi2020trustworthy}.

We then validate the performance of HO-CMP by simulating multiple experimental settings, encompassing both linear and non-linear outcome specifications and various types of interference, such as synthetic random geometric networks and real-world networks. Specifically, we introduce a \emph{Non-LinearInMeans} outcome specification, where the spillover effect is non-monotone in the fraction of treated neighbors; as an example of a complex treatment effect structure, we demonstrate how HO-CMP successfully estimates the total treatment effect by effectively utilizing higher-order moments of unit outcomes.

Simulating the experiments also allows us to calculate the ground truth value of the TTE, which remains unknown in real experiments, enabling us to compare the performance of HO-CMP to the ground truth TTE. Additionally, we benchmark HO-CMP against standard approaches such as difference-in-means and Horvitz-Thompson estimators, a recent technique of \cite{cortez2022staggered}, and a first-order CMP estimation, like the one by \cite{shirani2024causal}. We emphasize that a large body of recent estimators, e.g., \cite{jia2024clustered}, requires knowledge of the interference network and is not applicable in our setting. The results showcase HO-CMP outperforming the benchmarks in estimating the TTE over time and its flexibility to cover different outcome specifications and interference structures.

\paragraph{Related causal inference literature.}
The primary objective of research on causal inference in the context of network interference is to estimate causal effects while relaxing SUTVA. For this purpose, various assumptions and methods have been proposed. 
We briefly discuss the predominant ones.

A common approach to relax SUTVA is
partial interference. Under this assumption, units are divided into disjoint clusters and interference is assumed only within the same cluster \citep{sobel2006randomized,rosenbaum2007interference,hudgens2008toward,tchetgen2012causal,liu2014large,kang2016peer,viviano2020policy,bhattacharya2020causal,qu2021efficient,auerbach2021local,candogan2023correlated,ugander2023randomized}.
When interference extends across clusters, standard estimators become biased. To address this, \cite{eckles2016design} propose a cluster-randomized approach that randomizes treatment assignment across clusters, reducing bias. However, it requires knowledge of the clusters.

The other assumption to replace SUTVA is the Neighborhood Interference Assumption (NIA). NIA states that outcomes are only influenced by the treatments of neighboring units in the network. This assumption is commonly imposed in the literature that relaxes the SUTVA \citep{sussman2017elements}.
Some recent studies combine the NIA with the availability of either a fully or partially observed interference structure \citep{leung2020treatment, viviano2020experimental, agarwal2022network, belloni2022neighborhood, li2022random}. Without prior knowledge of the interference structure, \cite{cortez2022staggered} consider low-degree polynomial interactions among units in the network. \cite{leung2022causal} also introduces a weaker version of the NIA, where the interference between two units located far away from each other is allowed to be nonzero, but negligible.

Another approach is to facilitate the estimation of causal effects by setting restrictions on the network structure \citep{chin2018central,jagadeesan2020designs,wang2020design,li2022network,agarwal2022network,jagadeesan2020designs,leung2022causal}. These restrictions include bounding the largest node degree of the interference graph, limiting the degree of the dependency graph, observing specific patterns in the network, locally constrained interference structures, and restricting the topology of the interference network.

Driven by applications in marketplace platforms and two-sided marketplaces, several recent works have examined specific interference patterns \citep{holtz2020reducing,wager2021experimenting,munro2021treatment,johari2022experimental,harshaw2022designEC,farias2022markovian,bright2022reducing,farias2023correcting}. For example, \cite{farias2022markovian} study experiments in Markovian systems where interference effects propagate through constraints like limited inventory.

From another perspective, most of the existing literature on network interference focuses on the case of single-time point observation \citep{hudgens2008toward,aronow2017estimating,basse2019randomization,jackson2020adjusting,savje2021average}.  
These studies have provided insightful results on spatial interference effects, but they often overlook temporal variations of the treatment effect. Recently, there has been a shift to consider settings with multiple-time observations \citep{li2022network,boyarsky2023modeling}. However, the problem of considering the dynamics of units' outcomes remains understudied \citep{arkhangelsky2023causal}.

\section{Setup and Foundation}\label{sec:setup}
Consider a system of $N$ units indexed by $i \in [N]:= \{1, \cdots, N\}$ subject to a randomized experiment. The units are observed over a time horizon of $T+1$ periods and for each $t \in \{0, 1, \ldots, T\}$, we let $\treatment{i}{t}$ denote the treatment status of unit~$i$ during time period~$t$. For simplicity, we consider a Bernoulli randomized design such that $\treatment{i}{t} \sim \text{Bernoulli}(\pi_t)$. That is, at time $t$ unit~$i$ receives the \emph{treatment} with a probability of $\pi_t$, corresponding to $\treatment{i}{t} = 1$. Otherwise, unit~$i$ belongs to the \emph{control} group and $\treatment{i}{t} = 0$. In this context, we collectively define $\bm{\pi} = (\pi_0, \pi_1, \ldots, \pi_T)$ as the \emph{experimental design}. Then, following the potential outcome framework \citep{imbens2015causal}, let $\outcome{i}{t}{}(\Mtreatment{}{})$ represent the potential outcome of unit~$i$ at time~$t$, where $\Mtreatment{}{}$ denotes the entire treatment allocation matrix, with $\treatment{i}{t}$ as the entry in row $t$ and column $i$.

Administering the treatment of unit~$i$ at time~$t$ according to $\Otreatment{i}{t}$ (as one realization of the random variable $\treatment{i}{t}$), we use $\MOtreatment$ (as one realization of $\Mtreatment{}{}$) to show the matrix that captures the treatments of all units throughout the experiment; accordingly, we let $\Ooutcome{i}{t}{} = \outcome{i}{t}{}(\Mtreatment{}{}=\MOtreatment)$ be the observed outcome of unit~$i$ at time~$t$ under the treatment assignment $\MOtreatment$:
\begin{align*}
    \MOtreatment =
    \begin{bmatrix}
        \Otreatment{1}{0} & \Otreatment{2}{0} & \ldots & \Otreatment{N}{0} \\
        \Otreatment{1}{1} & \Otreatment{2}{1} & \ldots & \Otreatment{N}{1} \\
        \vdots & \vdots & \ddots & \vdots \\
        \Otreatment{1}{T} & \Otreatment{2}{T} & \ldots & \Otreatment{N}{T}
    \end{bmatrix},
    \quad\quad\quad
    \MOoutcome =
    \begin{bmatrix}
        \Ooutcome{1}{0}{} & \Ooutcome{2}{0}{} & \ldots & \Ooutcome{N}{0}{} \\
        \Ooutcome{1}{1}{} & \Ooutcome{2}{1}{} & \ldots & \Ooutcome{N}{1}{} \\
        \vdots & \vdots & \ddots & \vdots \\
        \Ooutcome{1}{T}{} & \Ooutcome{2}{T}{} & \ldots & \Ooutcome{N}{T}{}
    \end{bmatrix}.
\end{align*}
Observing $(\MOtreatment, \MOoutcome)$, we are interested in estimating the TTE of the intervention, defined as below:
\begin{equation}
    \label{eq:TTE_def}
    \TTE_t = \lim_{N \rightarrow \infty} \frac{1}{N} \sum_{i = 1}^N \left[\outcome{i}{t}{\bm{1}} - \outcome{i}{t}{\bm{0}}\right],\quad
    t = 0,1,\ldots, T,
\end{equation}
where $\bm{1}$ and $\bm{0}$ are matrices of all $1$ and all $0$ of appropriate dimensions (in this case, $T+1$ by $N$). Intuitively, the TTE measures the average effect of changing the treatment for the entire population. This is a common estimand in the network interference literature and provides important insights into the efficacy of the treatment for decision-makers \citep{jia2024clustered,chen2024optimized,viviano2023causal,yu2022estimating,cortez2022staggered}.

Deriving a practical and efficient estimator for the TTE is challenging due to the fact that we can observe the population only under one treatment scenario \citep{holland1986statistics}. Indeed, in Eq.~\eqref{eq:TTE_def}, we can observe at most one of $\outcome{i}{t}{\bm{1}}$ or $\outcome{i}{t}{\bm{0}}$, and often, neither.\footnote{We can only observe $\MOoutcome$ for one of exponentially many realizations of $\MOtreatment$.} In the following sections, we address this challenge by proposing a new class of estimators grounded in the CMP framework. These estimators rely on the efficient use of experimental data, $\MOoutcome$ and $\MOtreatment$, yielding accurate causal estimation under unknown network interference.

\subsection{Potential outcome specification and state evolution of the experiment}

In this section we provide a summary of the outcome specification and results of \cite{shirani2024causal} that we utilize in the remaining. For $t = 0, 1, \ldots, T-1$, we let $\outcomeg{t}{}: \RR \times \RR^{T+1} \mapsto \RR$ be an unknown measurable function. We also use $\Vtreatment{i}{} = \big(\treatment{i}{0}, \ldots, \treatment{i}{T}\big)^\top$ to denote the treatment assignment of unit~$i$ during the experiment. Accordingly, the treatment allocation matrix $\Mtreatment{}{}$ is a $T+1$ by $N$ matrix with columns equal to $\Vtreatment{i}{}$. Given potential outcomes $\outcome{j}{t}{}(\Mtreatment{}{})$ at time $t$ and $j\in [N]$, their outcomes in time period $t+1$ are specified by
\begin{align}
    \label{eq:outcome_function}
    \outcome{i}{t+1}{}(\Mtreatment{}{}) =
    \sum_{j=1}^N
    \IMatGl{ij}{} \outcomeg{t}{}\left(\outcome{j}{t}{}(\Mtreatment{}{}) ,\Vtreatment{j}{}\right)
    +
    \noise{i}{t},
    \quad\quad\quad
    t=0,1,\ldots,T-1,
\end{align}
where $\IMatGl{ij}{}$ quantifies the impact of unit $j$ on unit $i$ at time $t$ and $\noise{i}{t}$ is a zero-mean Gaussian noise with a variance of $\sigma_e^2$, accounting for measurement errors. In addition, we let $\IMatG{} = [\IMatGl{ij}{}]_{i,j\in [N]}$ and refer to it as the \emph{interference matrix}. Then, according to Eq.~\eqref{eq:outcome_function}, the function $\outcomeg{t}{}$ captures the impact of past outcomes and treatment assignments of other units on the current outcome of unit~$i$.

Now, fixing $t$, we define
\begin{align}
    \label{eq:average of outcomes}
    \AVO{}{}{t}(\Mtreatment{}{}) := \lim_{N\rightarrow\infty} \frac{1}{N} \sum_{i=1}^N \outcome{i}{t}{}(\Mtreatment{}{}),
    \quad\quad
    \VVO{}{}{t}(\Mtreatment{}{})^2 := \lim_{N\rightarrow\infty} \frac{1}{N} \sum_{i=1}^N \outcome{i}{t}{}(\Mtreatment{}{})^2 -  \AVO{}{}{t}(\Mtreatment{}{})^2.
\end{align}
Then, as shown by \cite{shirani2024causal}, whenever the elements of the interference matrix $\IMatGl{ij}{}$ are i.i.d. Gaussian random variables with mean $\mu/N$ and variance $\sigma^2/N$, under mild moment conditions on initial values $\outcome{i}{0}{}$, we have
\begin{equation}
    \label{eq:state_evolution}
    \begin{aligned}
        \AVO{}{}{t+1}(\Mtreatment{}{}) &\eqas
        \mu \EE\left[
        \outcomeg{t}{}\big(\AVO{}{}{t}(\Mtreatment{}{}) + \VVO{}{}{t}(\Mtreatment{}{}) Z_t, \Vtreatment{}{}\big)
        \right],
        \\
        \VVO{}{}{t+1}(\Mtreatment{}{})^2 &\eqas
        \sigma^2 \EE\left[
        \outcomeg{t}{}\big(\AVO{}{}{t}(\Mtreatment{}{}) + \VVO{}{}{t}(\Mtreatment{}{}) Z_t, \Vtreatment{}{}\big)^2
        \right] + \sigma_e^2,
    \end{aligned}
\end{equation}
where $Z_t \sim \cN(0,1)$ is independent from $\Vtreatment{}{} \sim \text{Bernoulli}(\bm \pi)$ (that is, $\treatment{}{t} \sim \text{Bernoulli}(\pi_t)$ and $\Vtreatment{}{} = (\treatment{}{0}, \treatment{}{1}, \ldots, \treatment{}{T})^\top$) and the equalities hold almost surely. We note that the theory behind this result is rooted in the AMP literature, going back to \cite{bolthausen2014iterative,bayati2011dynamics}. However, as 
\cite{shirani2024causal} note, there is a major distinction between the AMP literature and the above setting: in the AMP literature, the matrix $\IMatG{}$ is observed, and the aim is to construct proper functions $\outcomeg{t}{}$ for a completely different objective, which is studying the high-dimensional asymptotics of first-order algorithms. However, in the current context, the matrix $\IMatG{}$ and functions $\outcomeg{t}{}$ are \emph{unknown} and the goal is to estimate them.

Considering Eq.~\eqref{eq:average of outcomes}, the equations in \eqref{eq:state_evolution} determine the dynamics of the sample mean and sample variance of unit outcomes over time in large sample asymptotics, and are denoted by the State Evolution (SE) equations of the experiment \citep{shirani2024causal}. In the next section, we present an efficient algorithm to learn the state evolution dynamics outlined in Eq.~\eqref{eq:state_evolution}. This method enables us to accurately estimate the TTE defined in Eq.~\eqref{eq:TTE_def} and its corresponding confidence interval.

\section{Algorithm}
\label{Sec:Algorithm}
In this section, we introduce \emph{Higher-order Causal Message-passing} (\textsc{HO-CMP}) for estimating the TTE over the entire time horizon of the experiment. Briefly speaking, HO-CMP directly estimates the update function in the state evolution equations \eqref{eq:state_evolution}, thereby estimating counterfactual quantities while accounting for the impact of unknown network interference. 
To this end, by Eqs.~\eqref{eq:TTE_def} and \eqref{eq:average of outcomes}, we rewrite the TTE as the difference of the sample means in the large limits:
$$
\TTE_t = \nu_t(\bm{1}) -  \nu_t(\bm{0}).
$$
That means the problem of estimating the TTE is equivalent to estimating $\nu_t(\bm{1})$ and $\nu_t(\bm{0})$ using the observed data, denoted by $(\MOtreatment, \MOoutcome)$. On the other hand, considering the state evolution equations in \eqref{eq:state_evolution}, the \emph{system state} at time $t+1$, denoted by $(\nu_{t+1}(\MOtreatment), \rho_{t+1}(\MOtreatment)^2)$, is a (nonlinear) function of the system state distribution at time $t$, characterized by $(\nu_{t}(\MOtreatment), \rho_{t}(\MOtreatment)^2)$ and $\Vtreatment{}{}$, encompassing the sample mean and variance of observed outcomes as well as the design of the experiment. However, because the exact functional form and parameters of equations in \eqref{eq:state_evolution} are unknown, one cannot directly apply the SE to track the evolution of states. Therefore, we propose to estimate the unknown update functions in SE equations, utilizing the observed data $(\MOtreatment, \MOoutcome)$. For this purpose, we fix the treatment assignment matrix $\MOtreatment$ and define
\begin{equation*}
    \begin{aligned}
        \hat{\nu}_t(\MOtreatment) &:= \frac{1}{N} \sum_{i=1}^N y_t^{i},
        \quad\quad
        \hat{\rho}_t(\MOtreatment) ^2:= \frac{1}{N} \sum_{i=1}^N \big(y_t^{i} - \hat{\nu}_t(\MOtreatment) \big)^2,
        \\
        \bar{w}_t &:= \frac{1}{N} \sum_{i=1}^N \Otreatment{i}{t},
        \quad\quad
        \Vec{\bar{w}} := \left(\bar{w}_0, \ldots, \bar{w}_T\right)^\top.
    \end{aligned}
\end{equation*}
In addition, let $\Vec{\phi} = (\phi_k)_{k\in[K]}$ be a prespecified vector of measurable feature functions of current estimates of the sample mean $\hat{\nu}_t(\MOtreatment)$, sample variance $\hat{\rho}_t(\MOtreatment)^2$, and the design $\MOtreatment$. We define $\bm{x}_t$ to represent the \emph{feature vector} as follows:
\begin{align*}
\bm{x}_t  = \Vec{\phi}\Big(\hat{\nu}_t(\MOtreatment), \hat{\rho}_t(\MOtreatment), \MOtreatment \Big) \coloneqq \Big[\phi_1 \Big(\hat{\nu}_t(\MOtreatment), \hat{\rho}_t(\MOtreatment), \MOtreatment \Big)\,, \dots,\, \phi_K \Big(\hat{\nu}_t(\MOtreatment), \hat{\rho}_t(\MOtreatment),\MOtreatment \Big)\Big]^\top.
\end{align*}
Then, we formally propose learning the mapping $f_{\bm{\theta}}(\cdot)$ defined by, 
\begin{align}
(\hat{\nu}_{t+1}(\MOtreatment), \hat{\rho}_{t+1}(\MOtreatment)^2) &= f_{\bm{\theta}}(\bm{x}_t)
\label{eq:mapping-of-se}
\end{align}
We summarize the method in Algorithm~\ref{alg:HO-CMP}. Note in our experiment design we begin with all units under control by setting $\pi_0 = 0$, meaning no units receive treatment in period $0$. Additionally, to avoid non-identifiability issues, the experiment requires at least two stages, which corresponds to having at least two distinct values in the set $\{\pi_1, \ldots, \pi_T\}$.

The proposed HO-CMP method encompasses a rich family of
estimators, offering flexibility through the selection of feature functions $\{\phi_k\}_{k \in [K]}$ and model $f_{\bm{\theta}}(\cdot)$. 
Specifically, incorporating proper feature (basis) functions, with examples shown in Table \ref{table:HO-CMPs}, facilitates the extraction of informative patterns for learning the unknown nonlinear dynamics of the system throughout the experiment. In practice, one could choose these basis functions based on heuristics, domain knowledge, and prior information about the dynamics.

Specifically, in this paper, we consider the following estimators, as summarized in Table~\ref{table:HO-CMPs}. 

\textsc{FO-CMP} (First-Order Causal Message-Passing): This corresponds to the simple setting where $\nu_{t+1}(\MOtreatment)$ is assumed to be a function of the previous sample mean $\nu_{t}(\MOtreatment)$, the sample mean of the current treatment $\bar{w}_{t+1}$, and an additional term to model the interaction of the dynamics and previous treatments $\nu_{t}(\MOtreatment) \bar{w}_{t}$. Consequently, this model is irrelevant of the variance $\rho_{t+1}(\MOtreatment)^2$. This is true when $g_t$ takes a simple nonlinear form $g_t(y_t, \vec{w}) = \alpha y_t + \beta w_{t+1} + \gamma y_t w_t$. We remark that \textsc{FO-CMP} essentially uses the first state evolution equation in \eqref{eq:state_evolution} and fails to extract informative signals from the second evolution equation.

\textsc{HO-CMP} (Higher-Order Causal Message-Passing): HO-CMP further introduces the second-order terms $(\bar{w}_{t+1})^2$ and $\hat{\rho}_t(\MOtreatment)^2$ to model the nonlinear treatment effects. It improves data efficiency by utilizing both state evolution equations. It also allows estimation of higher order terms in Taylor series of $g_t$.

While FO-CMP extends the estimation algorithm in \cite{shirani2024causal} to accommodate experiments with more than two stages, HO-CMP introduces a new dimension to the estimation problem by incorporating second-order terms. This inclusion enhances data utilization, resulting in higher estimation efficiency in HO-CMP compared to FO-CMP.

\begin{table}[t]
\centering
\caption{Two examples of feature functions}
\label{table:HO-CMPs}
\renewcommand{\arraystretch}{1.5}
\begin{tabular}{@{}ccc@{}} 
\toprule
Algorithms  & Feature functions $\{\phi_k(\hat{\nu}_t(\MOtreatment),\hat{\rho}_t(\MOtreatment)^2,\MOtreatment)\}_{k \in [K]}$ & $f_{\bm{\theta}}(\cdot)$  \\ \midrule
\textsc{FO-CMP}   & $\left\{\hat{\nu}_t(\MOtreatment),\bar{w}_{t+1},\hat{\nu}_t(\MOtreatment)\cdot \bar{w}_{t}\right\}$   & linear regression    \\
\textsc{HO-CMP}   & $\left\{\hat{\nu}_t(\MOtreatment),\bar{w}_{t+1},\hat{\nu}_t(\MOtreatment)\cdot \bar{w}_{t}, \hat{\rho}_t(\MOtreatment)^2,\bar{w}_{t+1}^2\right\}$   & linear regression  \\
\bottomrule
\end{tabular}
\end{table}

\LinesNotNumbered%
\SetAlgoNoLine%
\begin{algorithm}
\label{alg:HO-CMP}
\caption{Higer-Order Causal Message Passing (HO-CMP) }\label{alg:HO}
\KwData{Observed data $(\MOtreatment, \MOoutcome)$, feature functions $\Vec{\phi} = (\phi_k)_{k\in [K]}$, machine learning model $f_{\bm{\theta}}(\cdot)$ }

\nonl\textbf{Step 1: Data processing}

\pushline\pushline\nonl 

\For{$t \gets 0$ \textbf{to} $T$}{
$\hat{\nu}_t(\MOtreatment) \leftarrow \frac{1}{N} \sum_{i=1}^N y_t^{i}$,

 $ \hat{\rho}_t(\MOtreatment)^2\leftarrow \frac{1}{N} \sum_{i=1}^N (y_t^{i} - \hat{\nu}_t(\MOtreatment) )^2$,

 $\bm{x}_t \gets \vec{\phi}\left(\hat{\nu}_t(\MOtreatment),\hat{\rho}_t(\MOtreatment)^2,\MOtreatment \right)$

}

\popline\popline \nonl\textbf{Step 2: Model Estimation}

\pushline\pushline\nonl

Estimate $f_{\bm{\theta}}$ from data $\left\{\left(\bm{x}_{t},\left(\hat{\nu}_{t+1}(\MOtreatment), \hat{\rho}_{t+1}(\MOtreatment)^2\right)\right)\right\}_{t\in[T-1]}$, guided by \eqref{eq:mapping-of-se}.

\popline\popline \nonl\textbf{Step 3: Counterfactual Estimation}

\pushline\pushline\nonl
$ \hat{\nu}_{0}({\bm{0}}) \leftarrow \hat{\nu}_0(\MOtreatment)$, $\hat{\nu}_{0}({\bm{1}})\leftarrow \hat{\nu}_0(\MOtreatment)$, $ \hat{\rho}_0({\bm{0}})^2\leftarrow \hat{\rho}_t(\MOtreatment)^2$, $ \hat{\rho}_0({\bm{1}})^2\leftarrow \hat{\rho}_t(\MOtreatment)^2$, $\widehat{\operatorname{TTE}}_{0} \gets 0$

\For{$t \gets 0$ \textbf{to} $T-1$}{

Compute the features  and predict the counterfactuals

$\bm{x}_t (\bm{0})\gets  \vec{\phi}\left(\hat{\nu}_{t}({\bm{0}}) ,\hat{\rho}_t({\bm{0}})^2, \bm{0}\right) $, $\bm{x}_t (\bm{1})\gets  \vec{\phi}\left(\hat{\nu}_{t}({\bm{1}}) ,\hat{\rho}_t({\bm{1}})^2, \bm{1}\right) $



$\left(\hat{\nu}_{t+1}({\bm{0}}) , \hat{\rho}_{t+1}(\bm{0})^2\right) \gets  f_{\bm{\theta}}(\bm{x}_t (\bm{0}))$, $\left(\hat{\nu}_{t+1}({\bm{1}}) , \hat{\rho}_{t+1}(\bm{1})^2\right) \gets  f_{\bm{\theta}}(\bm{x}_t (\bm{1}))$






Estimate the TTE

$\widehat{\TTE}_{t+1} \gets \hat{\nu}_{t+1}({\bm{1}}) - \hat{\nu}_{t+1}({\bm{0}}) $

}

\popline\popline\nonl \KwResult{$\left\{\widehat{\TTE}_t\right\}_{t \in[T]}$}

\end{algorithm}

    

\section{Experiments}\label{sec:experiment}

In this section, we use synthetic experiments under simulated and real-world network interference patterns, to compare the performance of FO-CMP and HO-CMP estimators, outlined in Table~\ref{table:HO-CMPs} and Algorithm~\ref{alg:HO-CMP}, with several benchmarks. First, we introduce the experimental design, benchmark estimators, interference patterns, and outcome specifications.

\paragraph{Experimental design.} We primarily focus on the staggered rollout design with $L$ distinct treated probabilities, denoted by $\pi^{(1)}, \cdots, \pi^{(L)}$, where $\pi^{(\ell)}$ increases monotonically with $\ell \in \{1,\ldots,L\}$. In the first $T^{(1)}$ periods, $\pi^{(1)}\times 100\%$ of units are in the treatment group. From $T^{(1)}$ to $T^{(2)}$ periods, $\pi^{(2)}\times 100\%$ of units are in the treatment group, and so forth. In the staggered rollout design, once a unit is allocated to treatment, it remains in the treatment group until the experiment concludes \citep{xiong2019optimal}. In the appendix, we also consider the Bernoulli randomized design, where the treatment is re-randomized at every time period, allowing units to switch between the treatment and control groups throughout the experiment. We use two values of $T=40,\; 200$ and set $L=4$, with $(\pi^{(1)}, \pi^{(2)}, \pi^{(3)}, \pi^{(4)}) = (0.1, 0.2, 0.4, 0.5)$. In the appendix, we show the impact of increasing $L$ or the maximum treatment probability $\pi^{(L)}$.

\paragraph{Benchmark estimators.} We first present two benchmark estimators commonly used for treatment effect estimation, both in settings with and without network interference. The final estimator is designed specifically for settings with unknown network interference \citep{cortez2022staggered}.

The first benchmark estimator is the standard difference-in-means (DM) estimator given by
\[\widehat{\TTE}^\dm_{t} = \frac{\sum_{j=1}^N \Ooutcome{j}{t}{} \Otreatment{j}{t}{} }{\sum_{j=1}^N \Otreatment{j}{t}{} } - \frac{\sum_{j=1}^N \Ooutcome{j}{t}{} (1-\Otreatment{j}{t}{}) }{\sum_{j=1}^N (1-\Otreatment{j}{t}{}) } \, , \]
which is the difference in average outcomes between treated and control units at each time period $t$.

The second benchmark is the standard \citet{horvitz1952generalization} (HT) estimator given by
\[\widehat{\TTE}^\hoth_{t} = \frac{1}{N} \sum_{j=1}^N \left[\frac{ \Ooutcome{j}{t}{} \Otreatment{j}{t}{} }{\pi_t} - \frac{ \Ooutcome{j}{t}{} (1-\Otreatment{j}{t}{}) }{1 - \pi_t }\right]  \, , \]
which weights observed outcomes by the inverse propensity score (i.e., $1/\pi_t$ or $1/(1-\pi_t)$).

The third benchmark estimator is the polynomial interpolation estimator (PolyFit) introduced by \cite{cortez2022staggered}. 
PolyFit operates by obtaining estimates for the average of outcomes at equilibrium for $L$
treated probabilities $\pi^{(1)}, \cdots, \pi^{(L)}$, denoted by $\nueq(\pi^{(1)}),\ldots,\nueq(\pi^{(L)})$, then it utilizes Lagrange interpolation method and obtains a degree-$L$ polynomial approximation for the function $\nueq: [0,1]\to\mathbb{R}$ which can be used to estimate the equilibrium values under global control and treatment, $\hat{\nu}_{\mathrm{equil}}(0)$ and $\hat{\nu}_{\mathrm{equil}}(1)$. Finally, TTE is estimated by 
\[
\widehat{\TTE}^{\mathrm{polyfit}}_t = \hat{\nu}_{\mathrm{equil}}(1) - \hat{\nu}_{\mathrm{equil}}(0)\,.
\]
On the one hand, PolyFit does not need any knowledge of the interference network; however, it comes at the expense of having to grapple with two challenges. First, it may incur a high variance as $L$ increases due to fitting a high-degree polynomial. The second challenge is that it needs accurate estimates for each $\nueq(\pi^{(\ell)})$, which requires treating $\pi^{(\ell)}$ fraction of units for a long enough number of periods so that the outcomes reach an equilibrium. This can be achieved if the staggered roll-out design is performed over a long enough horizon $T$ with each $T^{(\ell)}$ sufficiently large, and then estimating each $\nueq(\pi^{(\ell)})$ by sample average of outcomes at time $T^{(\ell)}$. However, when such a lengthy experiment is not feasible, the estimates for $\nueq(\pi^{(\ell)})$ will be less accurate.

\paragraph{Interference networks.}
We consider two networks (graphs). The first graph is a simulated random geometric graph model, studied by \cite{leung2022causal}. The second graph is a social network of Twitch users \citep{rozemberczki2021twitch}. In either scenario, we denote the adjacency matrix of the graph by $E \in \{0,1\}^{N \times N}$. For any $i$ and $j$, $E_{ij}$ equals $1$ if $j$ is a neighbor of $i$ and $0$ otherwise.

\paragraph{Outcome generating processes.}
We consider two outcome specifications to model monotone and non-monotone interference patterns. Specifically, for both settings, 
we generate outcomes using the following specification:
\[\outcome{i}{t+1}{} = \alpha + \beta \frac{\sum_{j=1}^N E_{ij} \outcome{j}{t}{} }{\sum_{j=1}^N E_{ij}} + \delta \cdot g\left( \frac{\sum_{j=1}^N E_{ij} \treatment{j}{t+1}{} }{\sum_{j=1}^N E_{ij}} \right)  + \gamma \treatment{i}{t+1}{}  + \epsilon^i_{t+1} \,, 
\]
where in the first setting, $g(\cdot)$ is taken to be the identity function, i.e., $g(x) = x$ for any $x$. Therefore, $\outcome{i}{t+1}{}$ depends linearly on the fraction of treated neighbors, and we refer to this setting as the \emph{LinearInMeans} outcome setting. This setting is widely studied in the causal inference literature \citep{cai2015social,eckles2016design,leung2022causal}. 

\begin{figure}[t]
  \centering
  \subfloat[$\nu_{\mathrm{equil}}(\pi)$ vs $\pi$]{
      \includegraphics[width=0.3\textwidth]{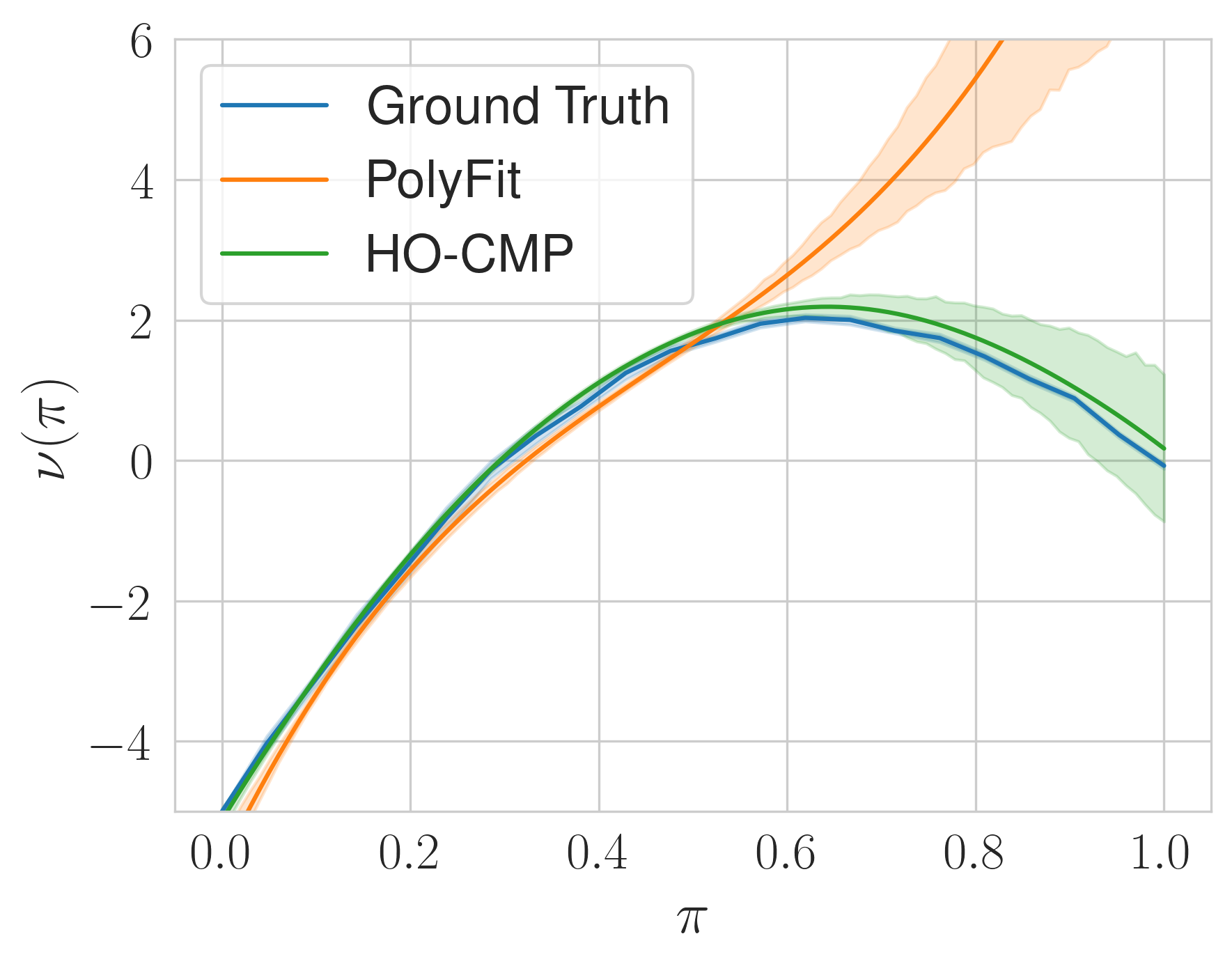}
  }
  \subfloat[Curve fitting of polyfit]{
      \includegraphics[width=0.3\textwidth]{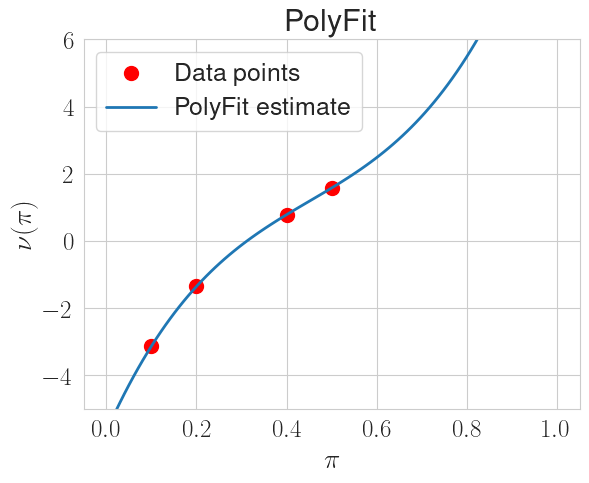}
  }
  \subfloat[Curve fitting of HO-CMP]{
      \includegraphics[width=0.3\textwidth]{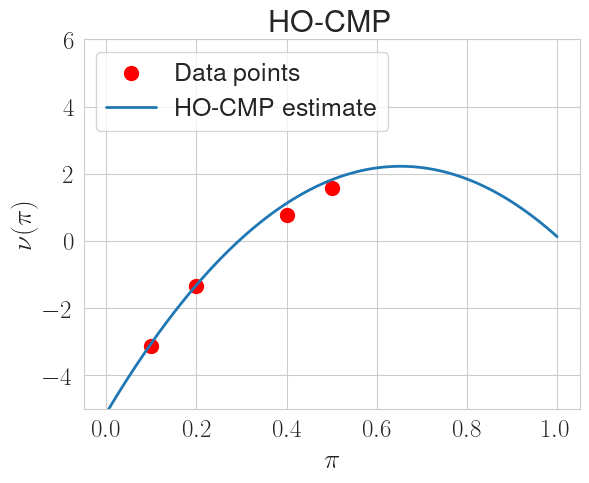}
  }
  
  \caption{(a) $\nu_{\mathrm{equil}}(\pi)$ with PolyFit and HO-CMP estimates across runs (Non-LinearInMeans). (b) and (c) show one sample estimates with observed data points.}
  \label{fig:nupi-visualization}
\end{figure}

In the second setting, $g(\cdot)$ is specified by a periodic function, i.e., $g(x) = \sin(\uppi x)$ for any $x$. Therefore, $\outcome{i}{t+1}{}$, on average, first increases and then decreases with the fraction of treated neighbors, as visualized by the Ground Truth curve in panel (a) of Figure \ref{fig:nupi-visualization}. We refer to this setting as the \emph{Non-LinearInMeans} outcome setting. 

\begin{figure}[t]
  \centering
  \subfloat[Random graph]{
      \includegraphics[width=0.5\textwidth]{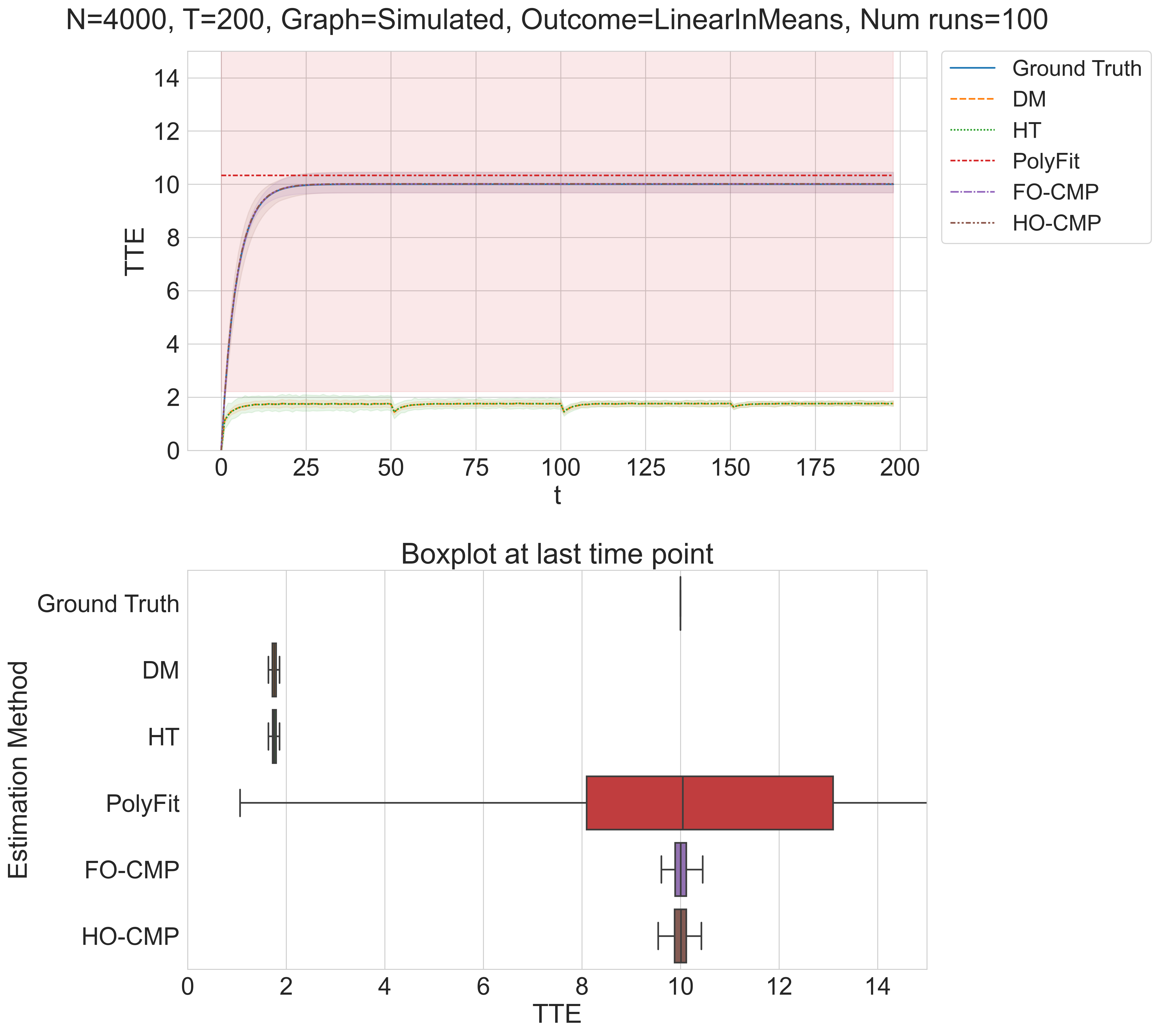}
  }
  \subfloat[Twitch graph]{
      \includegraphics[width=0.5\textwidth]{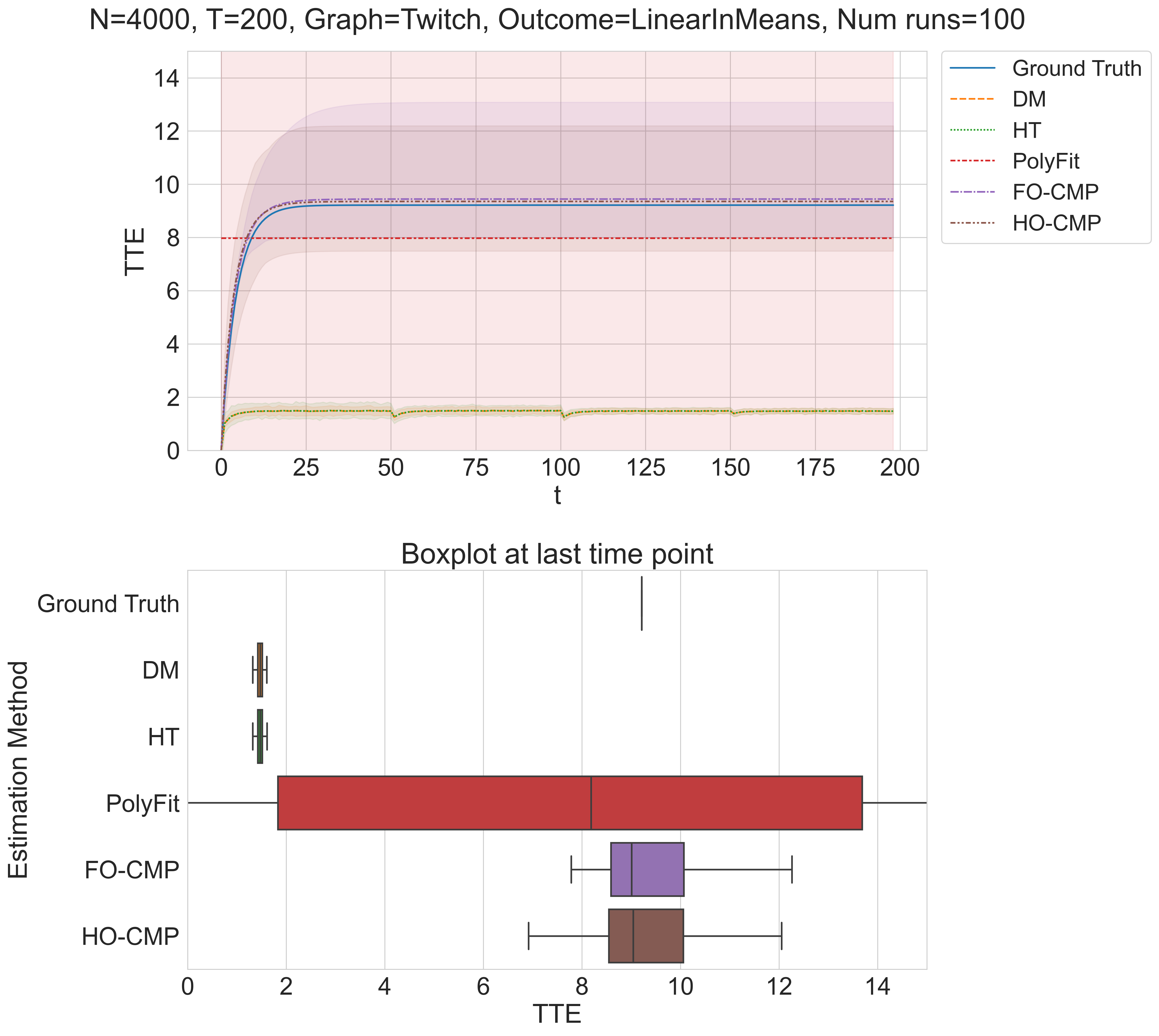}
  } 
  \caption{\emph{LinearInMeans with $T=200$.}
  $L=4$, with $(\pi^{(1)}, \pi^{(2)}, \pi^{(3)}, \pi^{(4)}) = (0.1, 0.2, 0.4, 0.5)$ and $T^{(\ell)} = 50 \ell$ for all $\ell \in \{1,2,3,4\}$. Shaded regions show 95\% percentile intervals of mean.
  }
  \label{fig:tte-linear-200}
\end{figure}

\begin{figure}[t]
  \centering
  \subfloat[Random graph]{
      \includegraphics[width=0.5\textwidth]{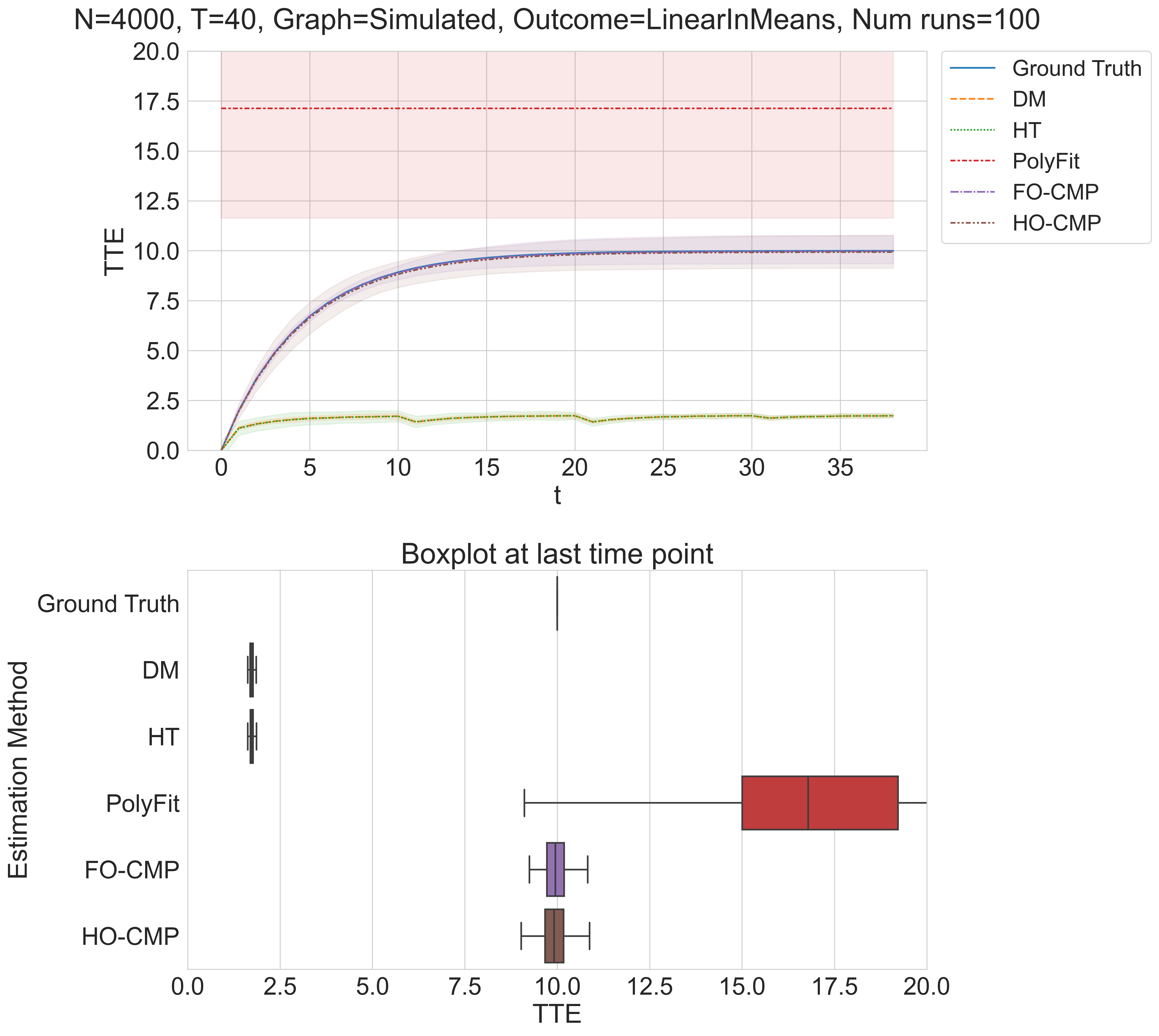}
  }
  \subfloat[Twitch graph]{
      \includegraphics[width=0.5\textwidth]{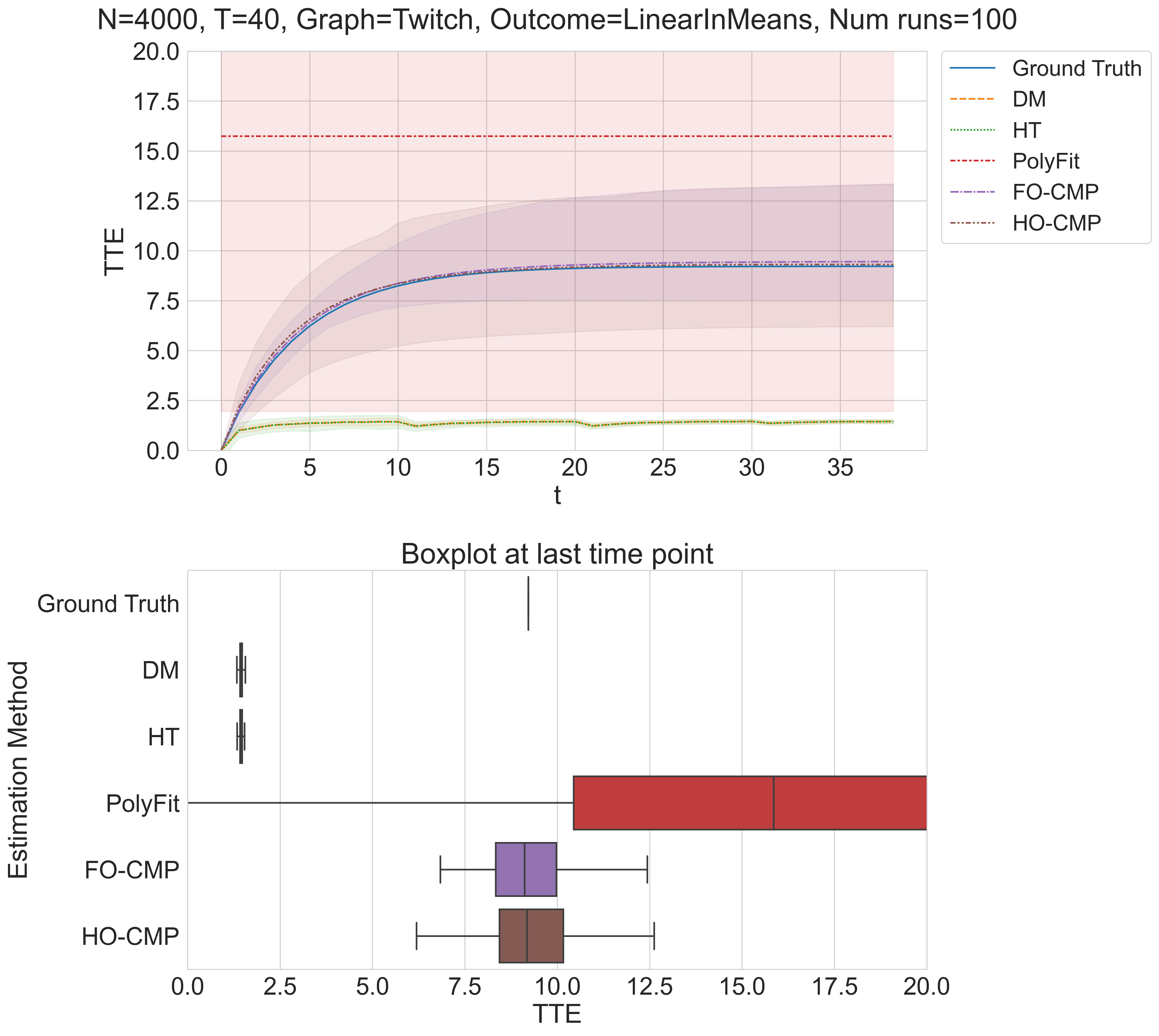}
  }
  \caption{\emph{LinearInMeans with $T=40$.}
  $L=4$, with $(\pi^{(1)}, \pi^{(2)}, \pi^{(3)}, \pi^{(4)}) = (0.1, 0.2, 0.4, 0.5)$ and $T^{(\ell)} = 10 \ell$ for all $\ell \in \{1,2,3,4\}$. Shaded regions show 95\% percentile intervals of mean.
  }
  \label{fig:tte-linear-40}
\end{figure}

\begin{figure}[t]
  \centering
  \subfloat[Random graph]{
      \includegraphics[width=0.5\textwidth]{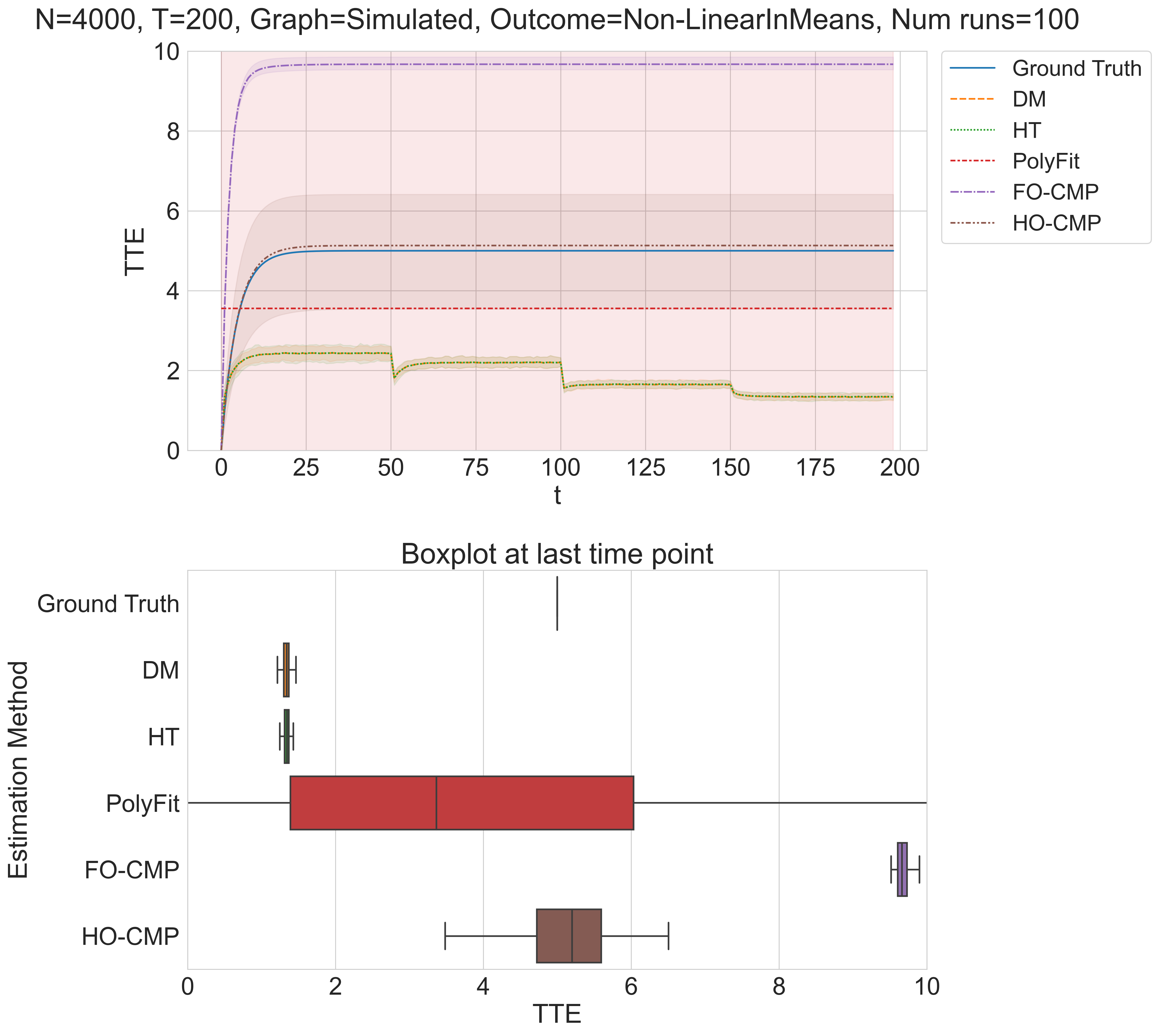}
  }
  \subfloat[Twitch graph]{
      \includegraphics[width=0.5\textwidth]{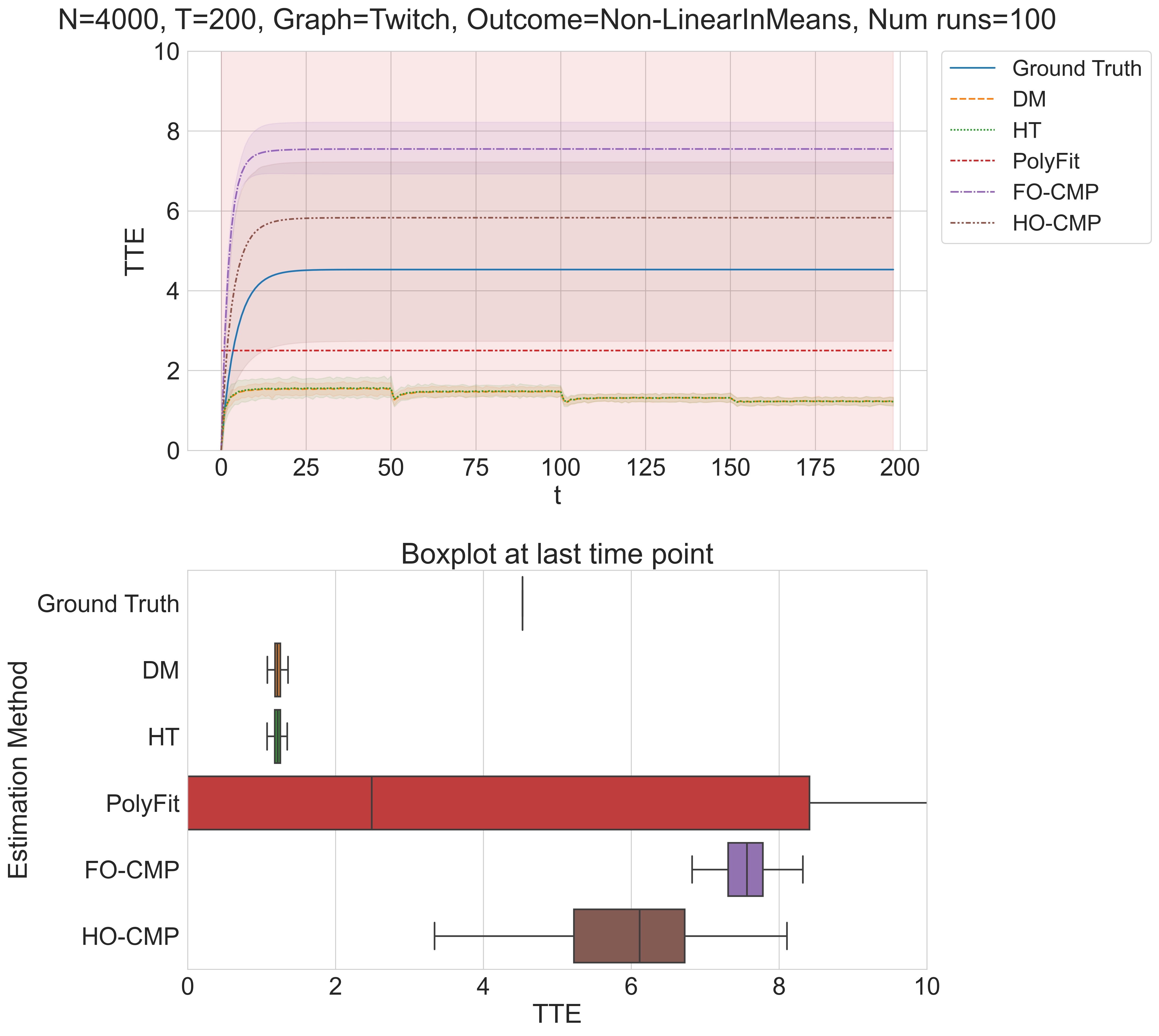}
  }
  \caption{\emph{Non-LinearInMeans with $T=200$.} $L=4$, with $(\pi^{(1)}, \pi^{(2)}, \pi^{(3)}, \pi^{(4)}) = (0.1, 0.2, 0.4, 0.5)$, and $T^{(\ell)} = 50 \ell$ for all $\ell$. Shaded regions show 95\% percentile intervals of mean.
  }
  \label{fig:tte-nonlinear-200}
\end{figure}

\begin{figure}[t]
  \centering
  \subfloat[Random graph]{
      \includegraphics[width=0.5\textwidth]{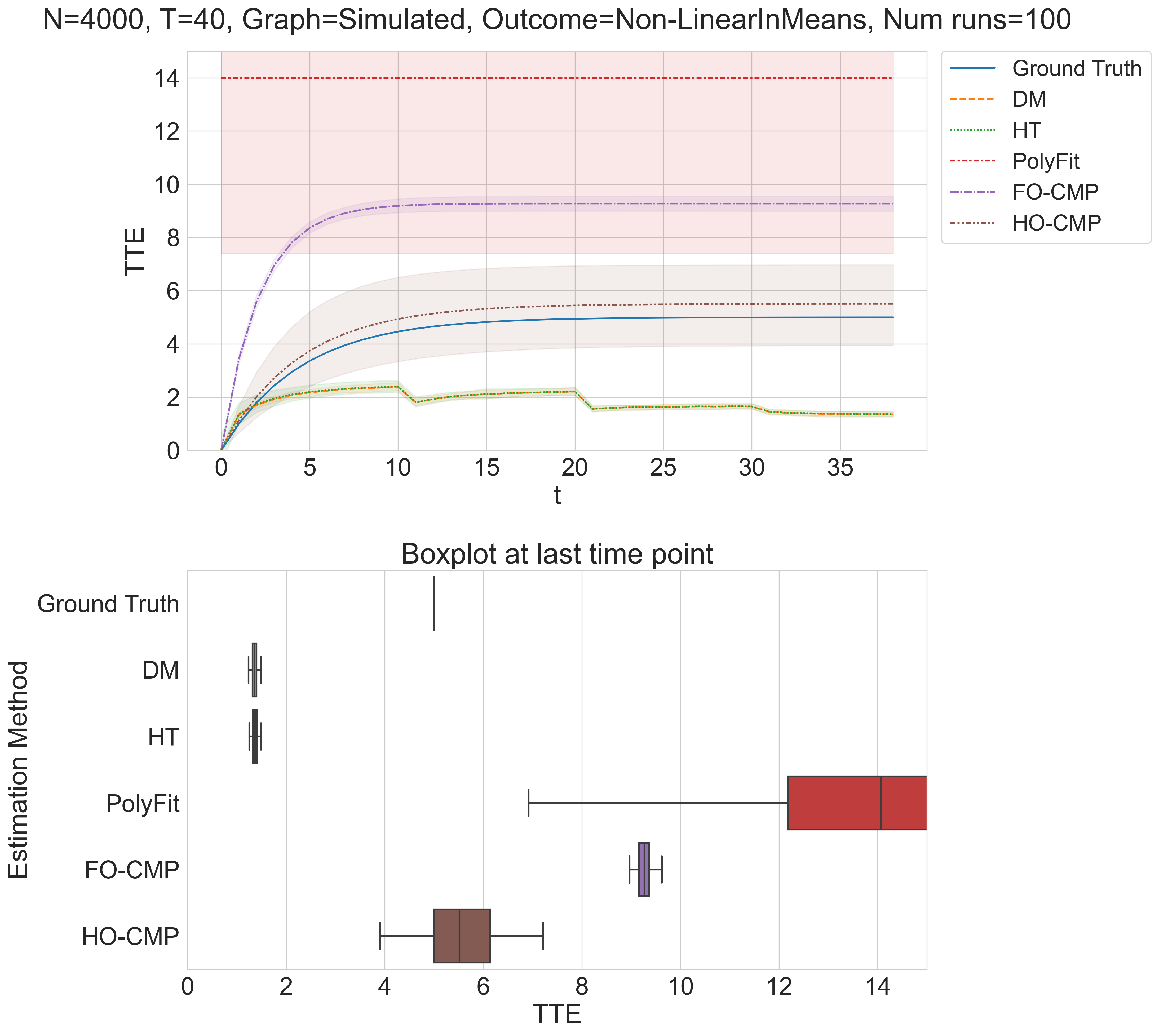}
  }
  \subfloat[Twitch graph]{\label{fig:TTE_Twitch_NonLinear_T_40}
      \includegraphics[width=0.5\textwidth]{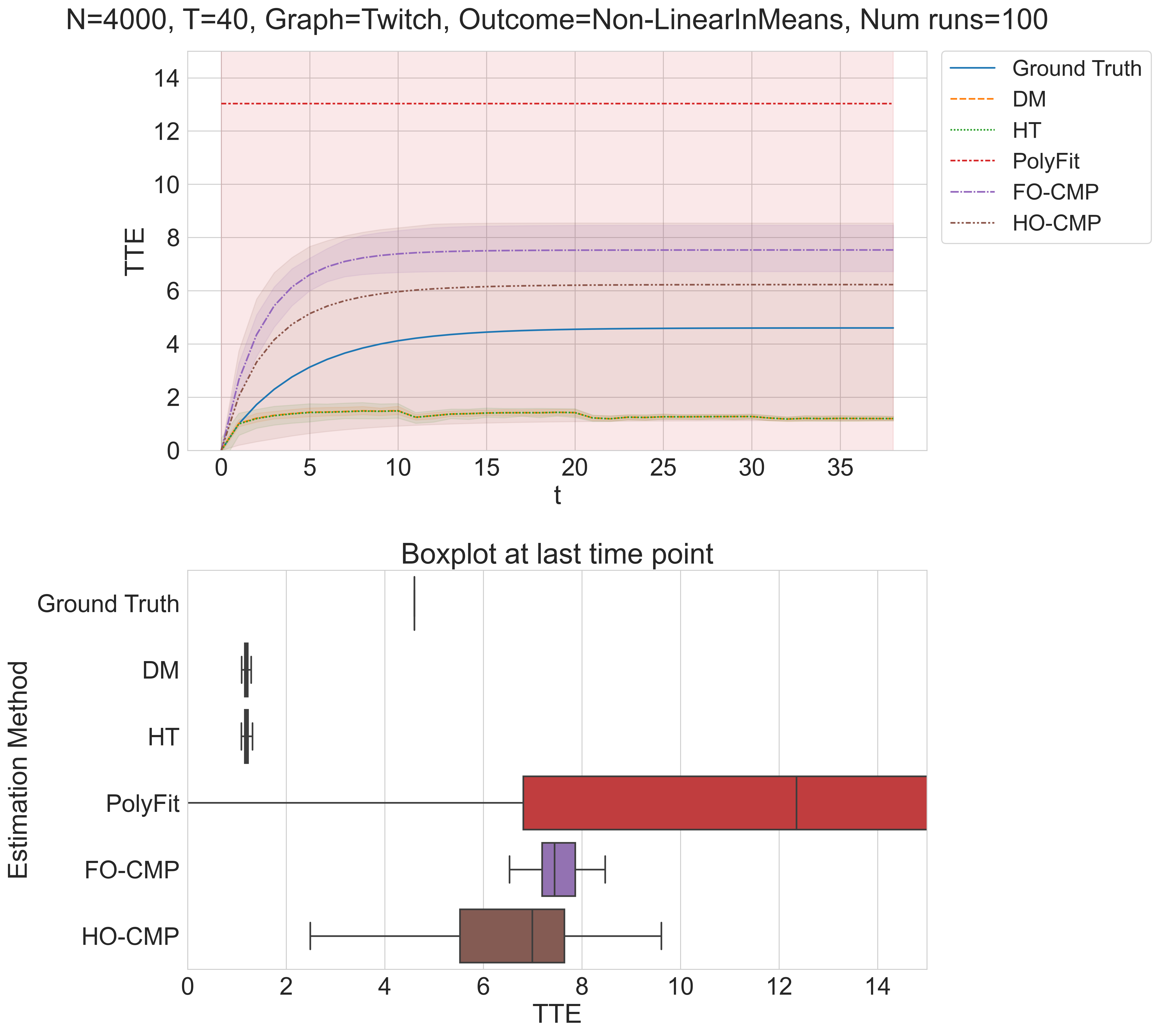}
  }
  \caption{\emph{Non-LinearInMeans with $T=40$.} $L=4$, with $(\pi^{(1)}, \pi^{(2)}, \pi^{(3)}, \pi^{(4)}) = (0.1, 0.2, 0.4, 0.5)$, and $T^{(\ell)} = 10 \ell$ for all $\ell$. Shaded regions show 95\% percentile intervals of mean.
  }
  \label{fig:tte-nonlinear-40}
\end{figure}

\paragraph{Results.} We compare FO-CMP and HO-CMP with the three benchmarks for estimating the TTE across the aforementioned outcome specifications and interference networks for long ($T=200$) and short ($T=40$) horizons. In each scenario, we perform 100 simulations of the synthetic experiment. The resulting distributions of ground truth and estimated TTEs are shown in Figures \ref{fig:tte-linear-200}-\ref{fig:tte-nonlinear-40}. All experiments were conducted on a MacBook Air with an Apple M1 chip and 16 GB of memory, with each setting taking about 15 minutes for 100 iterations. The key takeaways are as follows.

First, the DM and HT estimators exhibit significant bias across all cases. This is intuitive, as they estimate the TTE without accounting for the network interference.

Second, in the \emph{LinearInMeans} outcome setting, FO-CMP and HO-CMP achieve low estimation error and minimal bias. This holds for both long experiment durations ($T=200$), where outcomes reach equilibrium, and short experiment durations ($T=40$), where outcomes have not yet reached equilibrium, as shown in Figures \ref{fig:tte-linear-200} and \ref{fig:tte-linear-40}, respectively. 

Third, as expected, PolyFit's dependence on accurate estimates for each $\nueq(\pi^{(\ell)})$ requires a large $T$ to reduce estimation bias. This is evident when comparing Figures \ref{fig:tte-linear-200} and \ref{fig:tte-linear-40}: with a smaller $T$, PolyFit shows bias. This is also demonstrated in panel (b) of Figure \ref{fig:nupi-visualization}, where the red points—which represent sample averages of outcomes at $T^{(1)},\ldots,T^{(4)}$—have not yet converged and are slightly lower than their ground truth (equilibrium) values. This causes PolyFit's estimation of $\hat{\nu}_{\mathrm{equil}}(1)$ to be inaccurate, leading to a large bias. In contrast, HO-CMP, as shown in panel (c) of Figure \ref{fig:nupi-visualization}, is immune to this problem as it is designed to work with off-equilibrium data. Even with a larger $T$, the degree-$L$ polynomial estimation costs PolyFit with higher variance than both FO-CMP and HO-CMP, as shown in Figure \ref{fig:tte-linear-200}. Overall, this underscores the more efficient data utilization of FO-CMP and HO-CMP through their ability to leverage off-equilibrium data.

Fourth, in the \emph{Non-LinearInMeans} outcome setting, \textsc{HO-CMP} achieves substantially lower estimation error compared to \textsc{FO-CMP}, as shown in Figures \ref{fig:tte-nonlinear-200} and \ref{fig:tte-nonlinear-40}. This makes intuitive sense, as the higher-order terms in \textsc{HO-CMP} better capture the nonlinearity of $\nu_{\mathrm{equil}}(\pi)$ in $\pi$, while leveraging the additional data on sample variance dynamics over time, thereby enhancing the estimation accuracy.

Finally, the proposed estimation method demonstrates robustness across different experimental setups, including both \emph{LinearInMeans} and \emph{Non-LinearInMeans} outcome specifications. Additionally, robustness to graph structure---random versus Twitch graph---is evident from comparing the left and right plots in Figures \ref{fig:tte-linear-200}-\ref{fig:tte-nonlinear-40}. In Figure \ref{fig:variants_of_TTE_Twitch_NonLinear_T_40} of Appendix~\ref{sec:supp-exp}, we also demonstrate the robustness of the proposed methods to various parameters: the number of treatment probabilities $L$, the maximum treatment probability $\pi^{(L)}$, and the choice of experimental design (staggered rollout versus Bernoulli randomization).

\section{Conclusion}\label{sec:conclusion}
Estimating causal effects under pervasive interference presents significant challenges \citep{sussman2017elements}. Building on the causal message-passing framework of \cite{shirani2024causal}, we incorporate higher-order moments of observed outcomes and treatment probabilities to estimate the total treatment effect, without requiring knowledge of the interference network. Our approach leverages machine learning techniques to extract informative patterns from these higher moments, enabling our estimator to capture complex counterfactual behaviors, including non-monotonic trends in outcome means relative to treatment proportions. While we demonstrate strong performance across various outcome specifications and network structures, the framework's applicability may be limited when multiple outcome observations are unavailable.

\clearpage

\bibliographystyle{plainnat}
\bibliography{citations}

\begin{thebibliography}{66}
\providecommand{\natexlab}[1]{#1}
\providecommand{\url}[1]{\texttt{#1}}
\expandafter\ifx\csname urlstyle\endcsname\relax
  \providecommand{\doi}[1]{doi: #1}\else
  \providecommand{\doi}{doi: \begingroup \urlstyle{rm}\Url}\fi

\bibitem[Agarwal et~al.(2022)Agarwal, Cen, Shah, and Yu]{agarwal2022network}
Anish Agarwal, Sarah Cen, Devavrat Shah, and Christina~Lee Yu.
\newblock Network synthetic interventions: A framework for panel data with network interference.
\newblock \emph{arXiv preprint arXiv:2210.11355}, 2022.

\bibitem[Arkhangelsky and Imbens(2023)]{arkhangelsky2023causal}
Dmitry Arkhangelsky and Guido Imbens.
\newblock Causal models for longitudinal and panel data: A survey.
\newblock Technical report, National Bureau of Economic Research, 2023.

\bibitem[Aronow and Samii(2017)]{aronow2017estimating}
Peter~M Aronow and Cyrus Samii.
\newblock Estimating average causal effects under general interference, with application to a social network experiment.
\newblock 2017.

\bibitem[Auerbach and Tabord-Meehan(2021)]{auerbach2021local}
Eric Auerbach and Max Tabord-Meehan.
\newblock The local approach to causal inference under network interference.
\newblock \emph{arXiv preprint arXiv:2105.03810}, 2021.

\bibitem[Basse and Airoldi(2018)]{basse2018limitations}
Guillaume~W Basse and Edoardo~M Airoldi.
\newblock Limitations of design-based causal inference and a/b testing under arbitrary and network interference.
\newblock \emph{Sociological Methodology}, 48\penalty0 (1):\penalty0 136--151, 2018.

\bibitem[Basse et~al.(2019)Basse, Feller, and Toulis]{basse2019randomization}
Guillaume~W Basse, Avi Feller, and Panos Toulis.
\newblock Randomization tests of causal effects under interference.
\newblock \emph{Biometrika}, 106\penalty0 (2):\penalty0 487--494, 2019.

\bibitem[Bayati and Montanari(2011)]{bayati2011dynamics}
Mohsen Bayati and Andrea Montanari.
\newblock The dynamics of message passing on dense graphs, with applications to compressed sensing.
\newblock \emph{IEEE Transactions on Information Theory}, 57\penalty0 (2):\penalty0 764--785, 2011.

\bibitem[Belloni et~al.(2022)Belloni, Fang, and Volfovsky]{belloni2022neighborhood}
Alexandre Belloni, Fei Fang, and Alexander Volfovsky.
\newblock Neighborhood adaptive estimators for causal inference under network interference.
\newblock \emph{arXiv preprint arXiv:2212.03683}, 2022.

\bibitem[Bhattacharya et~al.(2020)Bhattacharya, Malinsky, and Shpitser]{bhattacharya2020causal}
Rohit Bhattacharya, Daniel Malinsky, and Ilya Shpitser.
\newblock Causal inference under interference and network uncertainty.
\newblock In \emph{Uncertainty in Artificial Intelligence}, pages 1028--1038. PMLR, 2020.

\bibitem[Blake and Coey(2014)]{blake2014why}
Thomas Blake and Dominic Coey.
\newblock Why marketplace experimentation is harder than it seems: the role of test-control interference.
\newblock In \emph{Proceedings of the Fifteenth ACM Conference on Economics and Computation}, EC '14, page 567–582, New York, NY, USA, 2014. Association for Computing Machinery.
\newblock ISBN 9781450325653.
\newblock \doi{10.1145/2600057.2602837}.
\newblock URL \url{https://doi.org/10.1145/2600057.2602837}.

\bibitem[Bolthausen(2014)]{bolthausen2014iterative}
Erwin Bolthausen.
\newblock An iterative construction of solutions of the tap equations for the sherrington--kirkpatrick model.
\newblock \emph{Communications in Mathematical Physics}, 325\penalty0 (1):\penalty0 333--366, 2014.

\bibitem[Bond et~al.(2012)Bond, Fariss, Jones, Kramer, Marlow, Settle, and Fowler]{bond201261}
Robert~M Bond, Christopher~J Fariss, Jason~J Jones, Adam~DI Kramer, Cameron Marlow, Jaime~E Settle, and James~H Fowler.
\newblock A 61-million-person experiment in social influence and political mobilization.
\newblock \emph{Nature}, 489\penalty0 (7415):\penalty0 295--298, 2012.

\bibitem[Boyarsky et~al.(2023)Boyarsky, Namkoong, and Pouget-Abadie]{boyarsky2023modeling}
Ariel Boyarsky, Hongseok Namkoong, and Jean Pouget-Abadie.
\newblock Modeling interference using experiment roll-out.
\newblock \emph{arXiv preprint arXiv:2305.10728}, 2023.

\bibitem[Bright et~al.(2022)Bright, Delarue, and Lobel]{bright2022reducing}
Ido Bright, Arthur Delarue, and Ilan Lobel.
\newblock Reducing marketplace interference bias via shadow prices.
\newblock \emph{arXiv preprint arXiv:2205.02274}, 2022.

\bibitem[Cai et~al.(2015)Cai, Janvry, and Sadoulet]{cai2015social}
Jing Cai, Alain~De Janvry, and Elisabeth Sadoulet.
\newblock Social networks and the decision to insure.
\newblock \emph{American Economic Journal: Applied Economics}, 7\penalty0 (2):\penalty0 81--108, 2015.

\bibitem[Candogan et~al.(2023)Candogan, Chen, and Niazadeh]{candogan2023correlated}
Ozan Candogan, Chen Chen, and Rad Niazadeh.
\newblock Correlated cluster-based randomized experiments: Robust variance minimization.
\newblock \emph{Management Science}, 2023.

\bibitem[Chen et~al.(2024)Chen, Li, Deng, and Wang]{chen2024optimized}
Qianyi Chen, Bo~Li, Lu~Deng, and Yong Wang.
\newblock Optimized covariance design for ab test on social network under interference.
\newblock \emph{Advances in Neural Information Processing Systems}, 36, 2024.

\bibitem[Chin(2018)]{chin2018central}
Alex Chin.
\newblock Central limit theorems via stein's method for randomized experiments under interference.
\newblock \emph{arXiv preprint arXiv:1804.03105}, 2018.

\bibitem[Choi(2017)]{choi2017estimation}
David Choi.
\newblock Estimation of monotone treatment effects in network experiments.
\newblock \emph{Journal of the American Statistical Association}, 112\penalty0 (519):\penalty0 1147--1155, 2017.

\bibitem[Cortez et~al.(2022)Cortez, Eichhorn, and Yu]{cortez2022staggered}
Mayleen Cortez, Matthew Eichhorn, and Christina Yu.
\newblock Staggered rollout designs enable causal inference under interference without network knowledge.
\newblock In \emph{Advances in Neural Information Processing Systems}, 2022.

\bibitem[Cox(1958)]{cox1958planning}
David~Roxbee Cox.
\newblock Planning of experiments.
\newblock 1958.

\bibitem[Donoho et~al.(2009)Donoho, Maleki, and Montanari]{donoho2009message}
David~L Donoho, Arian Maleki, and Andrea Montanari.
\newblock Message-passing algorithms for compressed sensing.
\newblock \emph{Proceedings of the National Academy of Sciences}, 106\penalty0 (45):\penalty0 18914--18919, 2009.

\bibitem[Eckles et~al.(2016)Eckles, Karrer, and Ugander]{eckles2016design}
Dean Eckles, Brian Karrer, and Johan Ugander.
\newblock Design and analysis of experiments in networks: Reducing bias from interference.
\newblock \emph{Journal of Causal Inference}, 5\penalty0 (1):\penalty0 20150021, 2016.

\bibitem[Farias et~al.(2022)Farias, Li, Peng, and Zheng]{farias2022markovian}
Vivek Farias, Andrew Li, Tianyi Peng, and Andrew Zheng.
\newblock Markovian interference in experiments.
\newblock \emph{Advances in Neural Information Processing Systems}, 35:\penalty0 535--549, 2022.

\bibitem[Farias et~al.(2023)Farias, Li, Peng, Ren, Zhang, and Zheng]{farias2023correcting}
Vivek~F Farias, Hao Li, Tianyi Peng, Xinyuyang Ren, Huawei Zhang, and Andrew Zheng.
\newblock Correcting for interference in experiments: A case study at douyin.
\newblock \emph{arXiv preprint arXiv:2305.02542}, 2023.

\bibitem[Forastiere et~al.(2022)Forastiere, Mealli, Wu, and Airoldi]{forastiere2022estimating}
Laura Forastiere, Fabrizia Mealli, Albert Wu, and Edoardo~M Airoldi.
\newblock Estimating causal effects under network interference with bayesian generalized propensity scores.
\newblock \emph{Journal of Machine Learning Research}, 23\penalty0 (289):\penalty0 1--61, 2022.

\bibitem[Harshaw et~al.(2022)Harshaw, S\"{a}vje, Eisenstat, Mirrokni, and Pouget-Abadie]{harshaw2022designEC}
Christopher Harshaw, Fredrik S\"{a}vje, David Eisenstat, Vahab Mirrokni, and Jean Pouget-Abadie.
\newblock Design and analysis of bipartite experiments under a linear exposure-response model.
\newblock In \emph{Proceedings of the 23rd ACM Conference on Economics and Computation}, EC '22, page 606, New York, NY, USA, 2022. Association for Computing Machinery.
\newblock ISBN 9781450391504.
\newblock URL \url{https://doi.org/10.1145/3490486.3538269}.

\bibitem[Holland(1986)]{holland1986statistics}
Paul~W Holland.
\newblock Statistics and causal inference.
\newblock \emph{Journal of the American statistical Association}, 81\penalty0 (396):\penalty0 945--960, 1986.

\bibitem[Holtz et~al.(2020)Holtz, Lobel, Liskovich, and Aral]{holtz2020reducing}
David Holtz, Ruben Lobel, Inessa Liskovich, and Sinan Aral.
\newblock Reducing interference bias in online marketplace pricing experiments.
\newblock \emph{arXiv preprint arXiv:2004.12489}, 2020.

\bibitem[Horvitz and Thompson(1952)]{horvitz1952generalization}
Daniel~G Horvitz and Donovan~J Thompson.
\newblock A generalization of sampling without replacement from a finite universe.
\newblock \emph{Journal of the American statistical Association}, pages 663--685, 1952.

\bibitem[Hudgens and Halloran(2012)]{hudgens2008toward}
Michael~G Hudgens and M~Elizabeth Halloran.
\newblock Toward causal inference with interference.
\newblock \emph{Journal of the American Statistical Association}, 103\penalty0 (482):\penalty0 832--842, 2012.

\bibitem[Imbens and Rubin(2015)]{imbens2015causal}
Guido~W Imbens and Donald~B Rubin.
\newblock \emph{Causal inference in statistics, social, and biomedical sciences}.
\newblock Cambridge University Press, 2015.

\bibitem[Jackson et~al.(2020)Jackson, Lin, and Yu]{jackson2020adjusting}
Matthew~O Jackson, Zhongjian Lin, and Ning~Neil Yu.
\newblock Adjusting for peer-influence in propensity scoring when estimating treatment effects.
\newblock \emph{Available at SSRN 3522256}, 2020.

\bibitem[Jagadeesan et~al.(2020)Jagadeesan, Pillai, and Volfovsky]{jagadeesan2020designs}
Ravi Jagadeesan, Natesh~S Pillai, and Alexander Volfovsky.
\newblock Designs for estimating the treatment effect in networks with interference.
\newblock 2020.

\bibitem[Jia et~al.(2024)Jia, Kallus, and Yu]{jia2024clustered}
Su~Jia, Nathan Kallus, and Christina~Lee Yu.
\newblock Clustered switchback experiments: Near-optimal rates under spatiotemporal interference, 2024.

\bibitem[Johari et~al.(2022)Johari, Li, Liskovich, and Weintraub]{johari2022experimental}
Ramesh Johari, Hannah Li, Inessa Liskovich, and Gabriel~Y Weintraub.
\newblock Experimental design in two-sided platforms: An analysis of bias.
\newblock \emph{Management Science}, 68\penalty0 (10):\penalty0 7069--7089, 2022.

\bibitem[Kang and Imbens(2016)]{kang2016peer}
Hyunseung Kang and Guido Imbens.
\newblock Peer encouragement designs in causal inference with partial interference and identification of local average network effects.
\newblock \emph{arXiv preprint arXiv:1609.04464}, 2016.

\bibitem[Karwa and Airoldi(2018)]{karwa2018systematic}
Vishesh Karwa and Edoardo~M Airoldi.
\newblock A systematic investigation of classical causal inference strategies under mis-specification due to network interference.
\newblock \emph{arXiv preprint arXiv:1810.08259}, 2018.

\bibitem[Kohavi et~al.(2020)Kohavi, Tang, and Xu]{kohavi2020trustworthy}
Ron Kohavi, Diane Tang, and Ya~Xu.
\newblock \emph{Trustworthy online controlled experiments: A practical guide to a/b testing}.
\newblock Cambridge University Press, 2020.

\bibitem[Leung(2020)]{leung2020treatment}
Michael~P Leung.
\newblock Treatment and spillover effects under network interference.
\newblock \emph{Review of Economics and Statistics}, 102\penalty0 (2):\penalty0 368--380, 2020.

\bibitem[Leung(2022)]{leung2022causal}
Michael~P Leung.
\newblock Causal inference under approximate neighborhood interference.
\newblock \emph{Econometrica}, 90\penalty0 (1):\penalty0 267--293, 2022.

\bibitem[Li and Wager(2022{\natexlab{a}})]{li2022network}
Shuangning Li and Stefan Wager.
\newblock Network interference in micro-randomized trials.
\newblock \emph{arXiv preprint arXiv:2202.05356}, 2022{\natexlab{a}}.

\bibitem[Li and Wager(2022{\natexlab{b}})]{li2022random}
Shuangning Li and Stefan Wager.
\newblock Random graph asymptotics for treatment effect estimation under network interference.
\newblock \emph{The Annals of Statistics}, 50\penalty0 (4):\penalty0 2334--2358, 2022{\natexlab{b}}.

\bibitem[Liu and Hudgens(2014)]{liu2014large}
Lan Liu and Michael~G Hudgens.
\newblock Large sample randomization inference of causal effects in the presence of interference.
\newblock \emph{Journal of the american statistical association}, 109\penalty0 (505):\penalty0 288--301, 2014.

\bibitem[Manski(1990)]{manski1990nonparametric}
Charles~F Manski.
\newblock Nonparametric bounds on treatment effects.
\newblock \emph{The American Economic Review}, 80\penalty0 (2):\penalty0 319--323, 1990.

\bibitem[Manski(2013)]{manski2013identification}
Charles~F Manski.
\newblock Identification of treatment response with social interactions.
\newblock \emph{The Econometrics Journal}, 16\penalty0 (1):\penalty0 S1--S23, 2013.

\bibitem[Mezard et~al.(1986)Mezard, Parisi, and Virasoro]{mezard1986spin}
M~Mezard, G~Parisi, and M~Virasoro.
\newblock \emph{Spin Glass Theory and Beyond, An Introduction to the Replica Method and Its Applications}.
\newblock World Scientific, Paris, Roma, November 1986.
\newblock \doi{10.1142/0271}.

\bibitem[Mezard and Montanari(2009)]{mezard2009information}
Marc Mezard and Andrea Montanari.
\newblock \emph{Information, physics, and computation}.
\newblock Oxford University Press, 2009.

\bibitem[Munro et~al.(2021)Munro, Wager, and Xu]{munro2021treatment}
Evan Munro, Stefan Wager, and Kuang Xu.
\newblock Treatment effects in market equilibrium.
\newblock \emph{arXiv preprint arXiv:2109.11647}, 2021.

\bibitem[Qu et~al.(2021)Qu, Xiong, Liu, and Imbens]{qu2021efficient}
Zhaonan Qu, Ruoxuan Xiong, Jizhou Liu, and Guido Imbens.
\newblock Efficient treatment effect estimation in observational studies under heterogeneous partial interference.
\newblock \emph{arXiv preprint arXiv:2107.12420}, 2021.

\bibitem[Rosenbaum(2007)]{rosenbaum2007interference}
Paul~R Rosenbaum.
\newblock Interference between units in randomized experiments.
\newblock \emph{Journal of the american statistical association}, 102\penalty0 (477):\penalty0 191--200, 2007.

\bibitem[Rozemberczki and Sarkar(2021)]{rozemberczki2021twitch}
Benedek Rozemberczki and Rik Sarkar.
\newblock Twitch gamers: a dataset for evaluating proximity preserving and structural role-based node embeddings, 2021.

\bibitem[Rubin(1978)]{rubin1978bayesian}
Donald~B Rubin.
\newblock Bayesian inference for causal effects: The role of randomization.
\newblock \emph{The Annals of statistics}, pages 34--58, 1978.

\bibitem[S{\"a}vje et~al.(2021)S{\"a}vje, Aronow, and Hudgens]{savje2021average}
Fredrik S{\"a}vje, Peter Aronow, and Michael Hudgens.
\newblock Average treatment effects in the presence of unknown interference.
\newblock \emph{Annals of statistics}, 49\penalty0 (2):\penalty0 673, 2021.

\bibitem[Shirani and Bayati(2024)]{shirani2024causal}
Sadegh Shirani and Mohsen Bayati.
\newblock Causal message-passing for experiments with unknown and general network interference.
\newblock \emph{Proceedings of the National Academy of Sciences}, 121\penalty0 (40):\penalty0 e2322232121, 2024.

\bibitem[Sobel(2006)]{sobel2006randomized}
Michael~E Sobel.
\newblock What do randomized studies of housing mobility demonstrate? causal inference in the face of interference.
\newblock \emph{Journal of the American Statistical Association}, 101\penalty0 (476):\penalty0 1398--1407, 2006.

\bibitem[Sussman and Airoldi(2017)]{sussman2017elements}
Daniel~L Sussman and Edoardo~M Airoldi.
\newblock Elements of estimation theory for causal effects in the presence of network interference.
\newblock \emph{arXiv preprint arXiv:1702.03578}, 2017.

\bibitem[Tchetgen and VanderWeele(2012)]{tchetgen2012causal}
Eric J~Tchetgen Tchetgen and Tyler~J VanderWeele.
\newblock On causal inference in the presence of interference.
\newblock \emph{Statistical methods in medical research}, 21\penalty0 (1):\penalty0 55--75, 2012.

\bibitem[Ugander and Yin(2023)]{ugander2023randomized}
Johan Ugander and Hao Yin.
\newblock Randomized graph cluster randomization.
\newblock \emph{Journal of Causal Inference}, 11\penalty0 (1):\penalty0 20220014, 2023.

\bibitem[Viviano(2020{\natexlab{a}})]{viviano2020experimental}
Davide Viviano.
\newblock Experimental design under network interference.
\newblock \emph{arXiv preprint arXiv:2003.08421}, 2020{\natexlab{a}}.

\bibitem[Viviano(2020{\natexlab{b}})]{viviano2020policy}
Davide Viviano.
\newblock Policy design in experiments with unknown interference.
\newblock \emph{arXiv preprint arXiv:2011.08174}, 2020{\natexlab{b}}.

\bibitem[Viviano et~al.(2023)Viviano, Lei, Imbens, Karrer, Schrijvers, and Shi]{viviano2023causal}
Davide Viviano, Lihua Lei, Guido Imbens, Brian Karrer, Okke Schrijvers, and Liang Shi.
\newblock Causal clustering: design of cluster experiments under network interference.
\newblock \emph{arXiv preprint arXiv:2310.14983}, 2023.

\bibitem[Wager and Xu(2021)]{wager2021experimenting}
Stefan Wager and Kuang Xu.
\newblock Experimenting in equilibrium.
\newblock \emph{Management Science}, 67\penalty0 (11):\penalty0 6694--6715, 2021.

\bibitem[Wang et~al.(2020)Wang, Samii, Chang, and Aronow]{wang2020design}
Ye~Wang, Cyrus Samii, Haoge Chang, and PM~Aronow.
\newblock Design-based inference for spatial experiments with interference.
\newblock \emph{arXiv preprint arXiv:2010.13599}, 2020.

\bibitem[Xiong et~al.(2024)Xiong, Athey, Bayati, and Imbens]{xiong2019optimal}
Ruoxuan Xiong, Susan Athey, Mohsen Bayati, and Guido Imbens.
\newblock Optimal experimental design for staggered rollouts.
\newblock \emph{Management Science}, 70\penalty0 (8):\penalty0 5317--5336, 2024.

\bibitem[Yu et~al.(2022)Yu, Airoldi, Borgs, and Chayes]{yu2022estimating}
Christina~Lee Yu, Edoardo~M Airoldi, Christian Borgs, and Jennifer~T Chayes.
\newblock Estimating the total treatment effect in randomized experiments with unknown network structure.
\newblock \emph{Proceedings of the National Academy of Sciences}, 119\penalty0 (44):\penalty0 e2208975119, 2022.

\end{thebibliography}

\appendix
\section{Supplementary Experiments}\label{sec:supp-exp}

\begin{figure}[h]
  \centering
  \subfloat[The same setting as Figure \ref{fig:TTE_Twitch_NonLinear_T_40} ]{
      \includegraphics[width=0.5\textwidth]{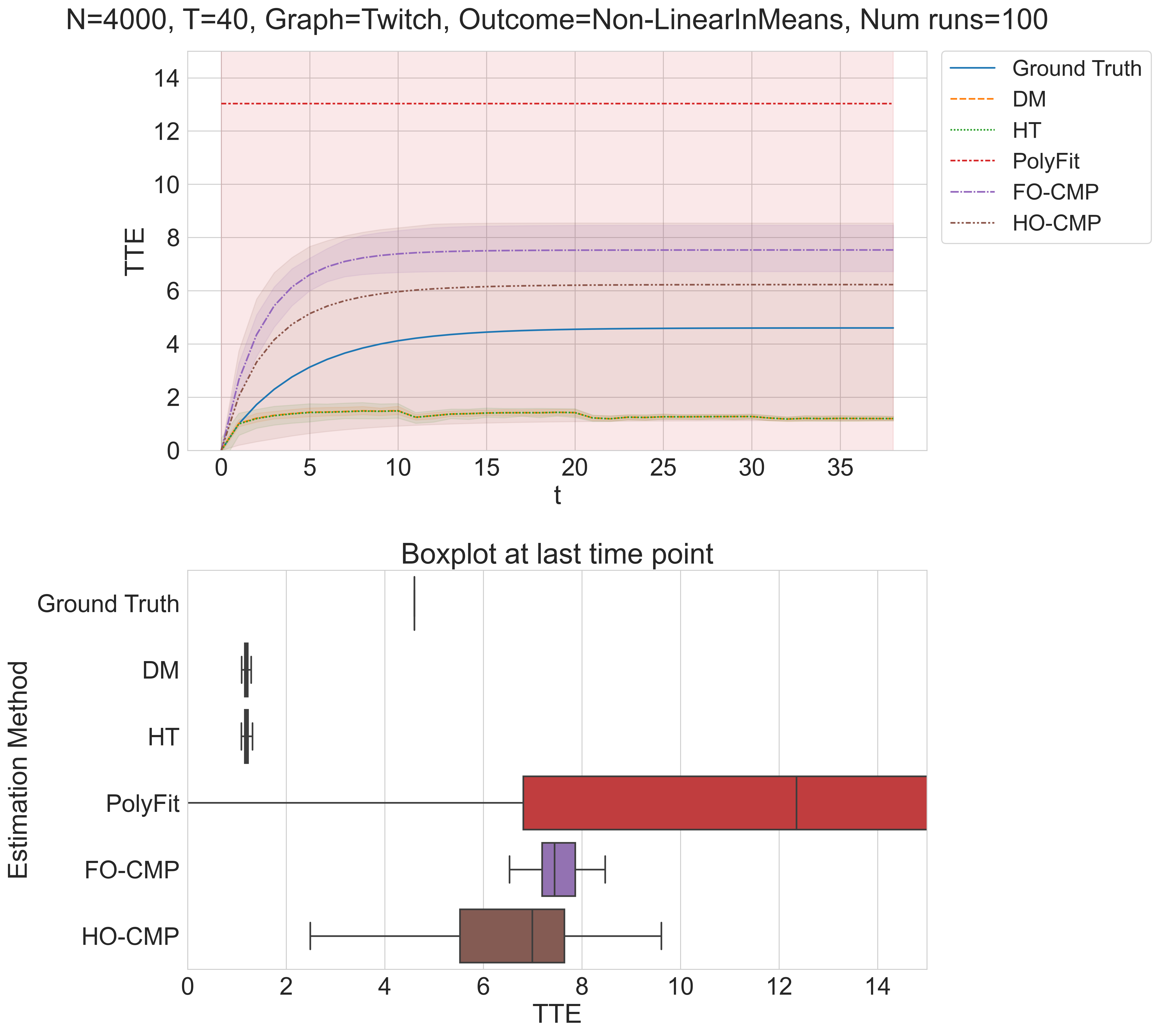}
 }
  \subfloat[Increasing number of $\pi$’s]{
      \includegraphics[width=0.5\textwidth]{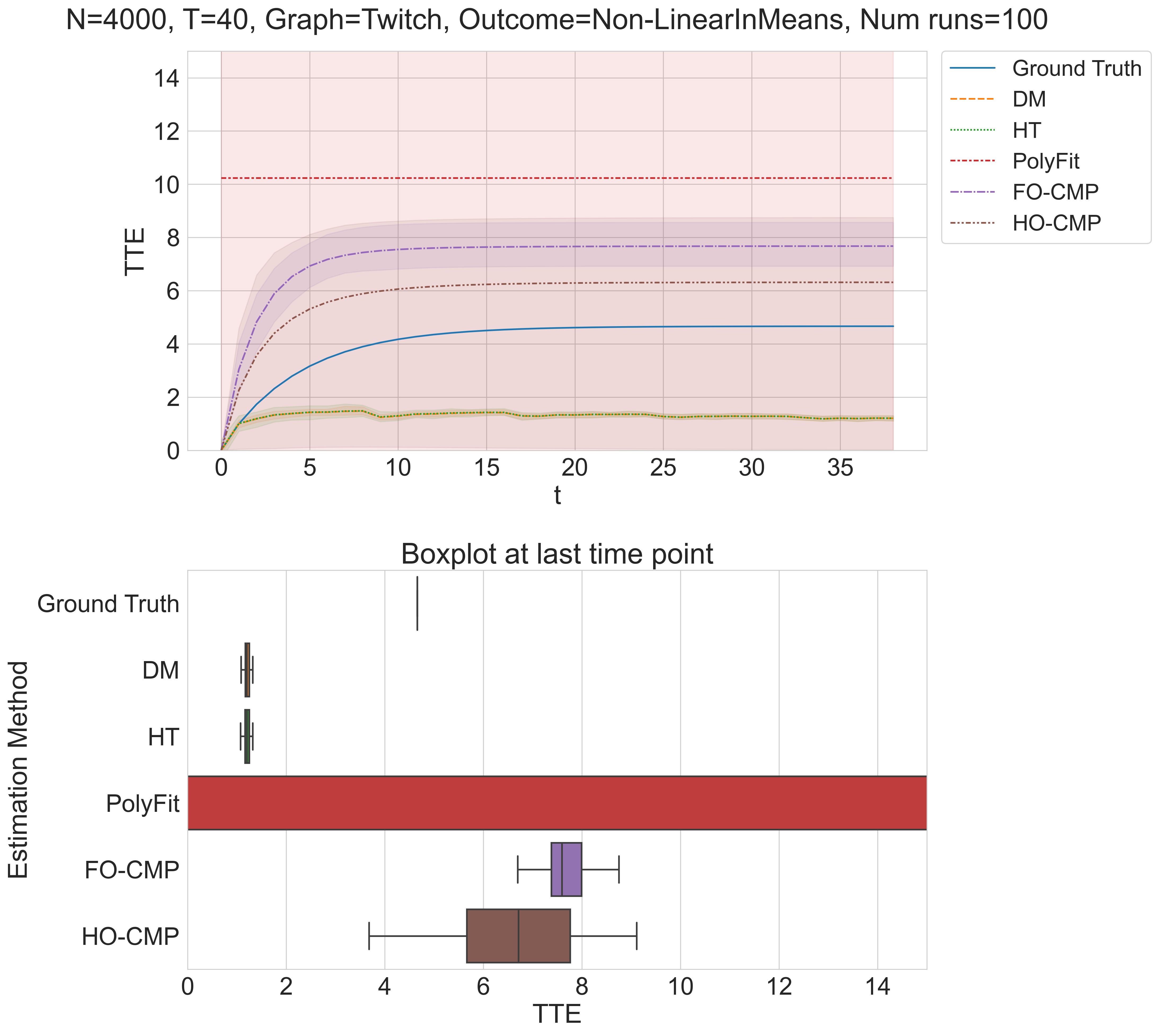}
  }\\
  \subfloat[Increasing the largest $\pi$]{
      \includegraphics[width=0.5\textwidth]{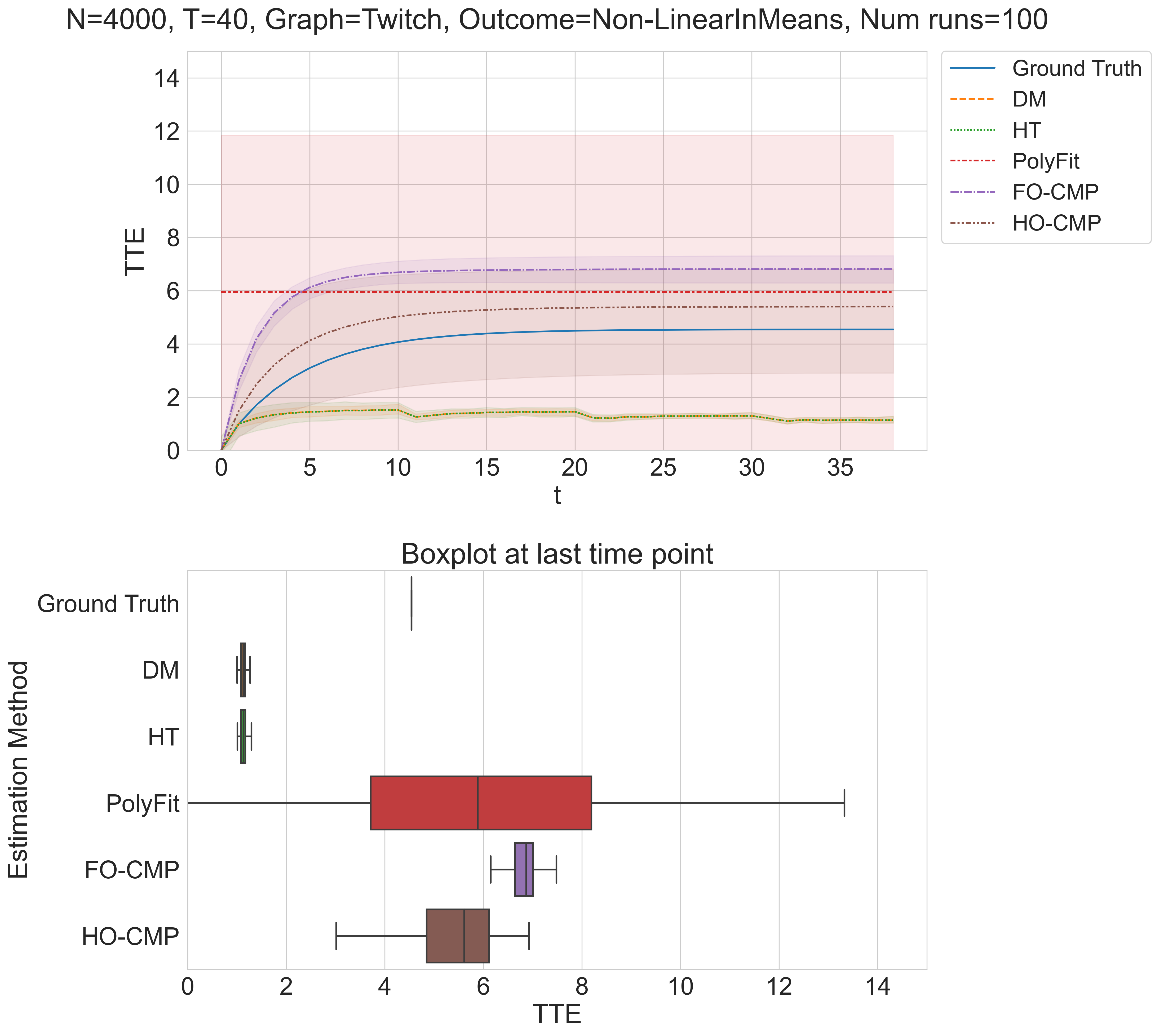}
  }
    \subfloat[Bernoulli randomized design]{
      \includegraphics[width=0.5\textwidth]{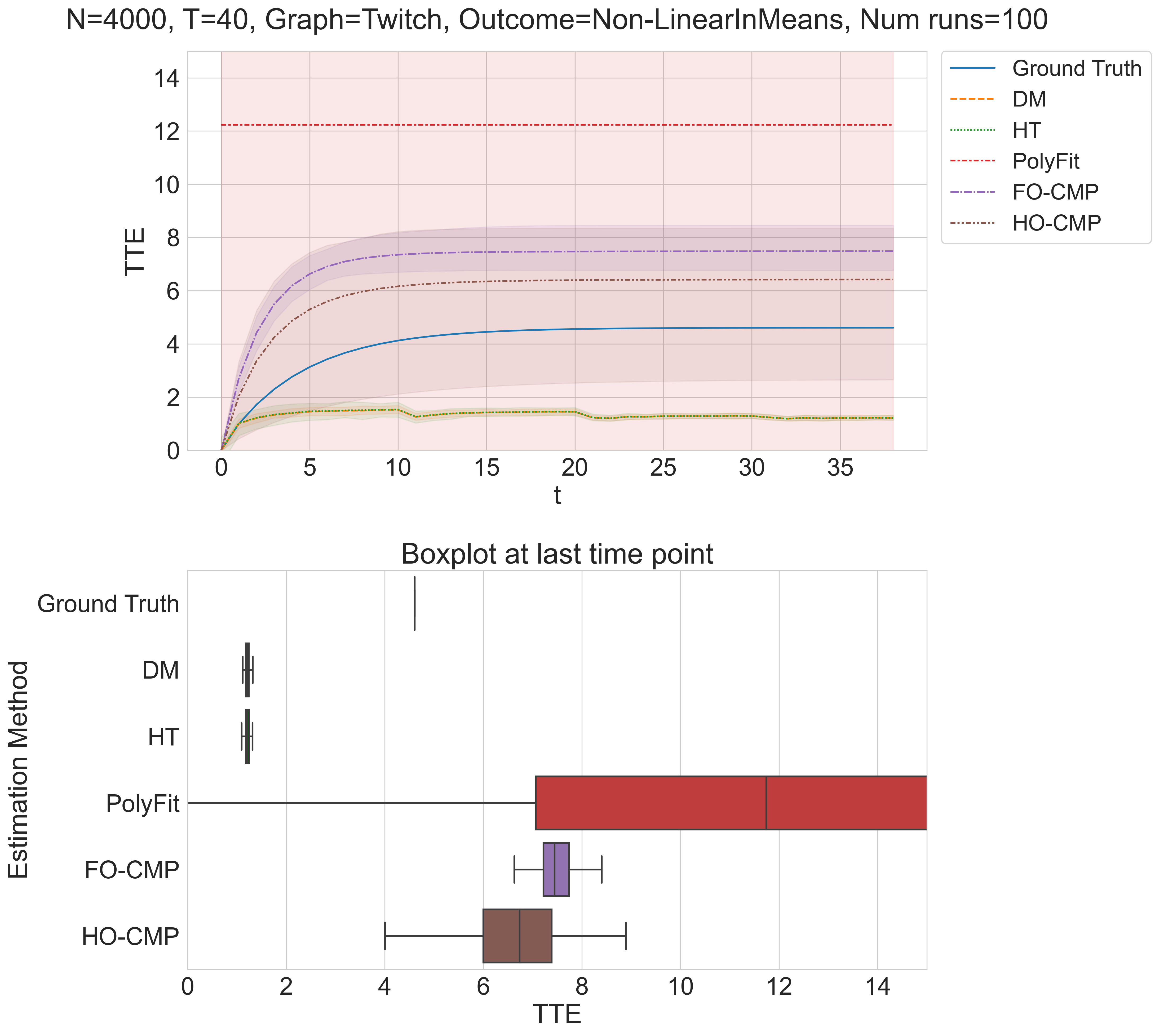}
  }
  \caption{\emph{Robustness check, under the Non-LinearInMeans with Twitch graph and $T=40$, i.e., setting of Figure \ref{fig:TTE_Twitch_NonLinear_T_40}.} 
  (a): Original Figure \ref{fig:TTE_Twitch_NonLinear_T_40}. (b): Increasing $L$: i.e., $\pi^{(\ell)} = 0.1 \ell$ and $T^{(\ell)} = 8\ell$ for all $\ell \in \{1,\ldots,5\}$.  (c): Increasing treatment probabilities, i.e., $(\pi^{(1)}, \pi^{(2)}, \pi^{(3)}, \pi^{(4)}) = (0.1, 0.2, 0.4, 0.6)$. (d): Using Bernoulli randomized design.
}
  \label{fig:variants_of_TTE_Twitch_NonLinear_T_40}
\end{figure}

\end{document}